\documentclass{article}

\PassOptionsToPackage{numbers, sort&compress}{natbib}

\usepackage[final]{neurips_2025}
\usepackage{subcaption} 

\usepackage[utf8]{inputenc} 
\usepackage[T1]{fontenc}    
\usepackage{hyperref}       
\usepackage{url}            
\usepackage{booktabs}       
\usepackage{amsfonts}       
\usepackage{nicefrac}       
\usepackage{microtype}      
\usepackage{xcolor}         
\usepackage{amsmath}
\usepackage{graphicx} 
\usepackage{soul}
\usepackage{wrapfig}
\usepackage{tikz}
\usepackage[normalem]{ulem}  

\newcommand{\PHENOMENON}{\textsc{Feedback Friction}}

\newcommand{\llamathree}{Llama-3.3-70B-Instruct}
\newcommand{\llamathreeshort}{Llama-3.3}
\newcommand{\llamafoursmall}{Llama-4-Scout-17B-16E-Instruct}
\newcommand{\llamafoursmallshort}{Llama-4-Scout}
\newcommand{\llamafourbig}{Llama-4-Maverick-17B-128E-Instruct-FP8}
\newcommand{\llamafourbigshort}{Llama-4-Maverick}
\newcommand{\claudeshort}{Claude 3.7}
\newcommand{\claudethinkshort}{Claude 3.7 Thinking}
\newcommand{\claude}{Claude 3.7 Sonnet}
\newcommand{\claudethinking}{Claude 3.7 Sonnet with Extended Thinking}

\newcommand{\gpt}{GPT-4.1 mini}
\newcommand{\feedbackone}{Binary Correctness Feedback ($F_1$)}
\newcommand{\feedbacktwo}{Self-Generated Reflective Feedback ($F_2$)}
\newcommand{\feedbackthree}{Strong-Model Reflective Feedback ($F_3$)}
\newcommand{\hyperfeedbackone}{\hyperlink{feedbackone}{Binary Correctness Feedback ($F_1$)}}
\newcommand{\hyperfeedbacktwo}{\hyperlink{feedbacktwo}{Self-Generated Reflective Feedback ($F_2$)}}
\newcommand{\hyperfeedbackthree}{\hyperlink{feedbackthree}{Strong-Model Reflective Feedback ($F_3$)}}

\title{\PHENOMENON{}: LLMs Struggle to\\Accurately Incorporate External Feedback}
\title{\PHENOMENON{}: Quantifying \\ LLMs' Struggle to Thoroughly Incorporate External Feedback}
\title{\PHENOMENON{}: Quantifying \\ LLMs' Resistance to External Feedback}
\title{\PHENOMENON{}: LLMs Struggle to \\ Fully Incorporate External Feedback}

\definecolor{mypurple}{RGB}{123,65, 212}

\author{%
  Dongwei Jiang$^{*\dagger}$ \quad Alvin Zhang$^{*\dagger}$ \quad Andrew Wang \quad Nicholas Andrews \quad Daniel Khashabi\\
  \texttt{djiang21@jhu.edu \quad bzhang90@jh.edu} \quad  \\
  Johns Hopkins University\\
  $^*$Equal contribution \quad $^{\dagger}$Corresponding authors
}

\begin{document}

\maketitle
\vspace{-0.5cm}  

\begin{abstract}
Recent studies have shown LLMs possess \textit{some} ability to improve their responses when given external feedback.
However, it remains unclear how effectively and thoroughly these models can incorporate extrinsic feedback. 
In an ideal scenario, if LLMs receive near-perfect and complete feedback, we would expect them to \emph{fully} integrate the feedback and reach correct solutions.
In this paper, we systematically investigate LLMs' ability to incorporate feedback by designing a controlled experimental environment. For each problem, a solver model attempts a solution, then a feedback generator with access to \textit{near-complete} ground-truth answers produces targeted feedback, 
after which the solver tries again.
We evaluate this pipeline across a diverse range of tasks, including math reasoning, knowledge reasoning, scientific reasoning, and general multi-domain evaluations with state-of-the-art language models including Claude 3.7 with extended thinking. Surprisingly, even under these near-ideal conditions, \textit{solver models consistently show resistance to feedback}, a limitation that we term \PHENOMENON.
To mitigate this limitation, we experiment with 
sampling-based strategies like progressive temperature increases and explicit rejection of previously attempted incorrect answers, which yield improvements but still fail to help models achieve target performance. 
We analyze \PHENOMENON{} and find that models' confidence on specific questions, measured by semantic entropy, predicts feedback resistance: high-confidence predictions remain resistant to external correction.
We hope that highlighting this issue in LLMs will help future research in self-improvement. 

\end{abstract}



\begin{figure}[hb]
    \centering
    \includegraphics[trim=0.0cm 1.9cm 2cm 1.5cm, scale=0.29]{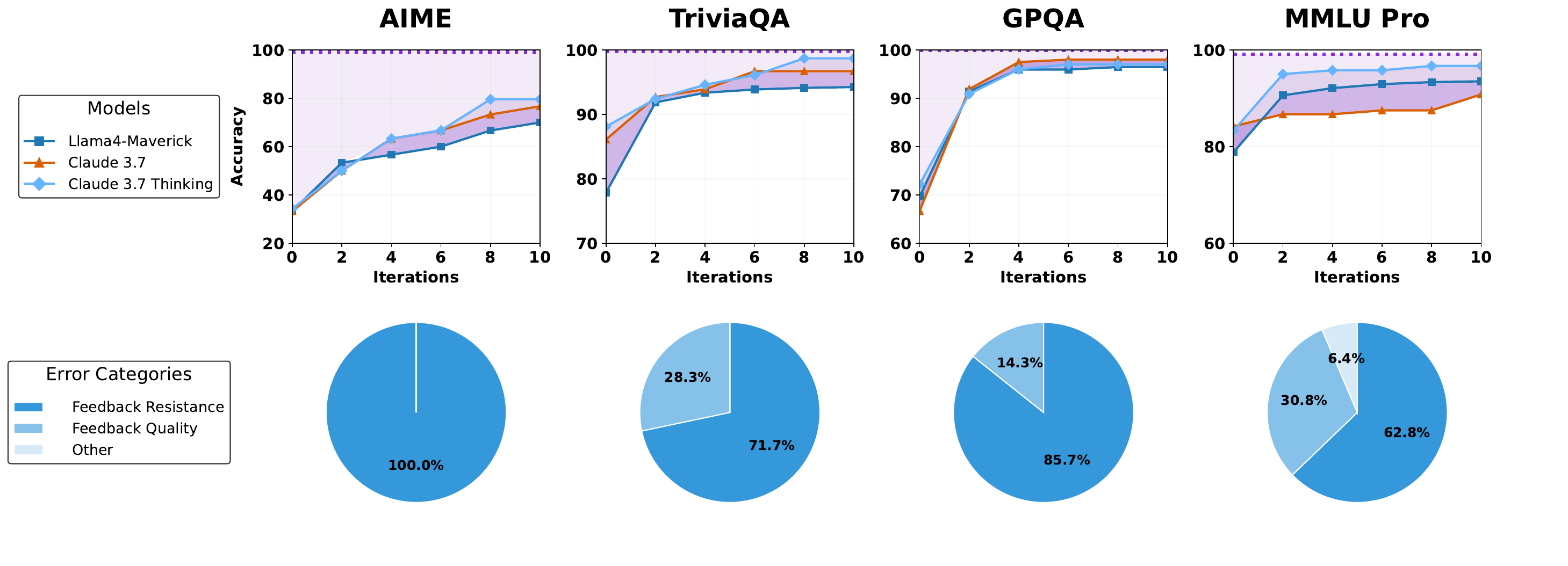} \\
    \caption{\textbf{Top:} 
    Accuracy of various \emph{solver} models when iteratively exposed to feedback from a \emph{feedback} model (GPT-4.1 mini) with access to ground-truth answers.
    The horizontal dotted line 
    \protect\tikz[baseline=-0.4ex]{\protect\draw[line width=1.5pt, dash pattern=on 1pt off 1pt, mypurple] (0,0) -- (0.33cm,0);}
    represents the target accuracy models could theoretically attain if they successfully incorporated all feedback 
    (details in \S\ref{sec:categorization}). 
    Despite receiving high-quality feedback, \textbf{solver models consistently plateau below their target accuracy}. \textbf{Bottom:} 
    Breakdown of questions that remained unsolved by the strongest solver model tested (\claudethinkshort{}) after multiple correction attempts.
    \textbf{Feedback resistance, rather than feedback quality issues, is responsible for the majority of persistent errors.}}
    \label{fig:teaser}
\end{figure}

\section{Introduction}

The prospect of self-improving Large Language Models (LLMs) has sparked both excitement and debate. Questions persist about LLMs' inherent ability to self-improve without external guidance~\citep{cant, jiang2024selfincorrectllmsstrugglediscriminating}, yet several studies have shown that LLMs can boost their performance when provided with accurate external feedback during test time without any parameter updates or training~\citep{Reflexion, wang2023voyager, can_correct}. However, while prior studies establish the \textit{existence} of performance gains from external feedback, \textit{the upper bounds} of such improvement remains largely unexplored. The extent of this improvement would have far-reaching implications for applications such as scientific discovery~\citep{si2024llmsgeneratenovelresearch, asai2024openscholarsynthesizingscientificliterature, lu2024aiscientistfullyautomated} and complex planning~\citep{zhou2023isrllmiterativeselfrefinedlarge, xie2024travelplannerbenchmarkrealworldplanning}, where repeated refinement cycles with feedback are critical for success. This raises a critical question: \textit{can models fully incorporate correct feedback to self-improve and reach their maximum potential?}

Answering this question is not straightforward, as self-improvement performance depends on two connected factors: feedback quality and the ability to incorporate feedback.
To decouple these two factors, we create a controlled experimental environment  that provides near-optimal conditions for feedback incorporation---one where models receive high-quality, targeted guidance based on complete ground-truth information.  \autoref{fig:self_improvement_loop} demonstrates our setup. The \emph{solver} model attempts to solve problems iteratively, receives feedback from a strong \emph{feedback} model on each incorrect answer, and retries again for up to 10 consecutive iterations. The feedback generator has access to the complete history of all previous attempts and responses, enabling targeted guidance that addresses persistent errors.

While this controlled environment with high-quality feedback provides an ideal testing ground, the effectiveness of feedback incorporation may depend significantly on the \emph{quality} of the feedback. To assess them, we implement three increasingly sophisticated feedback mechanisms to determine how much feedback quality impacts model performance and whether any level of feedback enables models to reach target accuracy. \hyperfeedbackone{} provides simple indication of correctness (e.g., ``the answer is wrong''). \hyperfeedbacktwo{} has the model itself analyze potential errors using correct answers and available solution steps (e.g., ``you correctly set up the equation but made an error when computing the derivative...''). Finally, \hyperfeedbackthree{} uses a more capable external model (GPT-4.1 mini) to generate feedback. 
Details of our evaluation framework and feedback generation process can be found in \S\ref{sec:method}.

We conduct a systematic study using all three forms of feedback across strong frontier models including \llamathree{}, \llamafoursmall{}, \llamafourbig{}, \claude{}, and \claudethinking{}. We evaluate these models across diverse tasks, including AIME 2024, MATH-500, TriviaQA, PopQA, MMLU, MMLU Pro, GPQA, and two synthetic digit multiplication tasks. While higher-quality feedback does improve self-improvement performance, a fundamental limitation persists. As shown in \autoref{fig:teaser}, even with our best feedback mechanism---\hyperfeedbackthree{}---models consistently fall short of the target accuracy (i.e., the accuracy of an ideal model if it successfully incorporated all the given feedback).
We name this weakness \PHENOMENON{} (\S\ref{sec:results}).

Our analysis in \S\ref{sec:analysis} reveals several key insights into \PHENOMENON{}. First, we categorize errors that persist after multiple feedback iterations and find that \textit{feedback resistance} (i.e., models failing to incorporate clear and accurate feedback) \textit{is the dominant failure mode across all tasks} (\S\ref{sec:categorization}). Second,
we attempt to mitigate it by avoiding repeated wrong answers through sampling strategies. 
While performance improves across tasks, models still fall below their target accuracy (\S\ref{sec:sampling}). 
Finally, we investigate why models resist feedback (\S\ref{sec:understanding}).
Our investigations rule out several apparent causes, including reasoning complexity, data familiarity, and whether specific questions consistently challenge all models. However, we uncover two key findings: First, using semantic entropy to measure confidence, we find that less confident models show greater relative improvements from feedback, indicating confident models are more resistant to correction. Second, models consistently claim to understand feedback and express willingness to update their beliefs (>95\%) yet fail to actually incorporate corrections—revealing a disconnect between stated intentions and actual behavior. In summary, having shown that state-of-the-art LLMs consistently resist external feedback, we identify important patterns in this resistance while ruling out several intuitive explanations.\footnote{Code for this work is available at: \url{https://github.com/JHU-CLSP/Feedback-Friction}}


\section{ A controlled framework to surface \PHENOMENON{}}\label{sec:method}
Our framework employs two key components: an iterative self-improvement loop that allows models multiple opportunities to correct their mistakes (\S\ref{sec:setup}), and a spectrum of feedback mechanisms with varying levels of detail and guidance (\S\ref{sec:feedback_mechanisms}).

\subsection{Setup for iterative self-improvement loop}
\label{sec:setup}

Given a task $T$ with evaluation dataset $D = \{(x_i, y_i)\}_{i=1}^{m}$ and evaluation metric $f$, we establish an iterative improvement protocol with two distinct models: a solver model $M_{\text{solver}}$ and a feedback generator model $M_{\text{feedback}}$.

For each input $x_i$, the solver model produces an initial answer $a_1(x_i)$ using the standard task prompt. We evaluate the correctness of this answer using $f(a_1(x_i), y_i)$, where $y_i$ is the ground truth. If the answer is incorrect, the feedback generator $M_{\text{feedback}}$ creates targeted guidance $g_1$ based on the current answer and ground-truth information.

For iteration $k \geq 1$, we construct the prompt $p_{k+1}(x_i) = \text{concat}(x_i, \text{history}_k)$, where $\text{history}_k$ contains all previous answers and feedback pairs $\{(a_1, g_1),(a_2, g_2), ..., (a_k, g_k)\}$.
This process repeats for up to $k=10$ iterations or until the correct answer is generated. In our empirical experiments, we observed that performance improvements tend to plateau within 10 iterations, suggesting a practical upper limit for the iterative refinement process.

The overall accuracy for the dataset at iteration $k$ is measured as the fraction of all problems solved correctly: $\text{Acc}_k = \frac{1}{m}\sum_{i=1}^m \mathbf{1}[f(a_k(x_i), y_i) = 1]$, where $\mathbf{1}[\cdot]$ is the indicator function.

\begin{figure*}[t!]
    \centering

    \includegraphics[trim=0.6cm -0.4cm 23cm 0.25cm,scale=0.157]{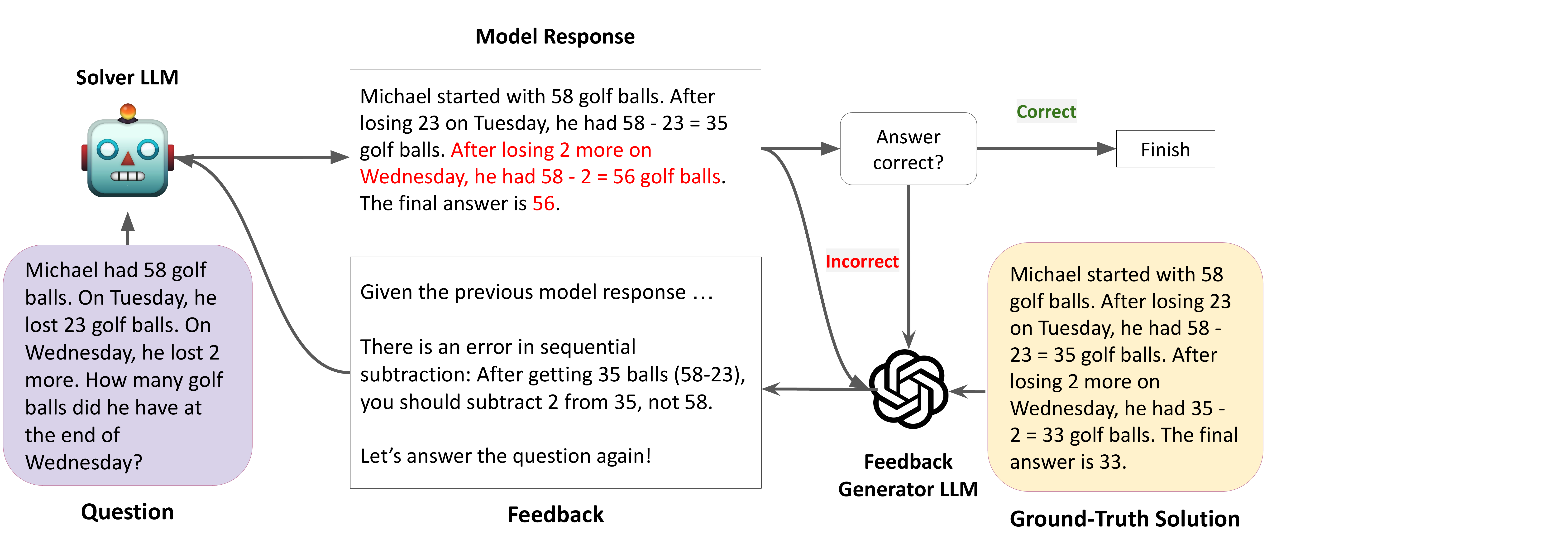}
\caption{\textbf{Iterative self-improvement loop.} The process involves: (1) a \textit{solver} model generating an answer, (2) a \textit{feedback} model generating feedback given incorrect responses and the ground-truth correct answer, and (3) the \textit{solver} attempting again with this feedback. This cycle repeats for up to 10
iterations or until a correct answer is produced.}
    \label{fig:self_improvement_loop}
\end{figure*}

By construction, since we iterate over the incorrect responses only,
the accuracy sequence $\{\text{Acc}_1, \text{Acc}_2, ..., \text{Acc}_K\}$ is monotonically non-decreasing, as we retain correct answers across iterations and only modify incorrect ones. This protocol, illustrated in \autoref{fig:self_improvement_loop}, provides a controlled environment for measuring how effectively models incorporate feedback while maintaining consistent evaluation criteria throughout the improvement process.

\subsection{Designing different feedback mechanisms for iterative self-improvement}
\label{sec:feedback_mechanisms}

We investigate three distinct feedback mechanisms for the self-improvement process, each offering progressively greater guidance and error specificity. All mechanisms are designed to identify errors without directly revealing the correct answer, ensuring a fair evaluation of the model's ability to incorporate feedback. 

\paragraph{\feedbackone}
\label{F1} 
\hypertarget{feedbackone}{The} simplest form provides only correctness information:
\begin{equation*}
F_{1}^{\text{binary}}(x_i, y_i) = \text{``The answer is wrong!''} \quad \text{if } f(a(x_i), y_i) = 0
\end{equation*}
where $a(x_i)$ is the solver model's answer and $f$ is the evaluation function.

\paragraph{\feedbacktwo} 
\label{F2} 
\hypertarget{feedbacktwo}{In} this approach, the solver model analyzes its own response using available information:
\begin{equation*}
F_{2}^\text{self}(x_i, y_i) = M_{\text{solver}}(\text{concat}(x_i, a(x_i), y_i, s_i, p_\text{prompt}))
\end{equation*}
where $p_\text{prompt}$ is the instruction: ``Please give me feedback on which solution step is wrong and how to get to the correct answer without revealing the answer.'' Here, $s_i$ represents the ground-truth solution process when available. For datasets without detailed solutions, only the answer $y_i$ is provided.

\paragraph{\feedbackthree} 
\label{F3} 
\hypertarget{feedbackthree}{We} employ a more capable external model with access to the same information:
\begin{equation*}
F_{3}^\text{strong}(x_i, y_i) = M_{\text{strong}}(\text{concat}(x_i, a(x_i), y_i, s_i, p_\text{prompt}))
\end{equation*}
where $M_{\text{strong}}$ represents a more powerful model than $M_{\text{solver}}$ in providing feedbacks. 

\section{Experimental results}\label{sec:exp_results}

In this section, we present comprehensive experimental results evaluating \PHENOMENON{}. We first describe our experimental setup, including tasks, prompts, inference setups, and model configurations (\S\ref{sec:exp_setup}). We then demonstrate how models consistently plateau below target performance regardless of the feedback mechanisms employed (\S\ref{sec:results}).

\subsection{Experimental setup} \label{sec:exp_setup}

\noindent\paragraph{Tasks and metrics.} We employ nine diverse tasks for evaluation, deliberately choosing objective tasks with clear ground-truth answers to ensure reliable evaluation of feedback incorporation. \footnote{Using another LLM to evaluate more subjective tasks like instruction following or translation could lead to issues like reward hacking and unreliable assessments.} Our tasks include: AIME 2024 \citep{huggingfaceh4_aime_2024} and MATH-500 \citep{hendrycks2021measuringmathematicalproblemsolving} for mathematical problem-solving, TriviaQA \citep{TriviaQA} and PopQA \citep{mallen2022trustlmretrieval} for knowledge reasoning, MMLU \citep{mmlu} and MMLU Pro \citep{wang2024mmluprorobustchallengingmultitask} for multi-domain evaluation, GPQA \citep{rein2023gpqagraduatelevelgoogleproofqa} for complex scientific reasoning, and two synthetic digit multiplication tasks. The first synthetic task involves 5-digit multiplication (e.g., 78934 × 62851), 
while the second task applies hexadecimal multiplication rules to decimal numbers (i.e., first mapping 0-9 to themselves and 10-15 to their hexadecimal digits ($10 \rightarrow A$, ..., $15 \rightarrow F$), then performing the arithmetic as if operating in base-16---creating a counterfactual \citep{wu2024reasoningrecitingexploringcapabilities} arithmetic setting that challenges models' learned numerical reasoning patterns. For both tasks, ground-truth solutions are generated using deterministic templates that break down the multiplication into smaller multiplication and addition operations. We include these synthetic tasks specifically to remove potential confounding variables from semantic context (more details about these tasks can be found in \autoref{appendix:synthetic_task}).

Other than the two synthetic tasks, only AIME 2024, GPQA, and MATH-500 include complete solutions, while the others provide only final answers. 
For MMLU and MMLU Pro, the multiple-choice format raises concerns about whether models might solve problems through simple elimination strategies rather than genuine reasoning (e.g., selecting A in the first iteration, B in the second, C in the third, etc). However, our analysis reveals that models exhibit surprising choice persistence even when provided with corrective feedback. Rather than systematically eliminating options or switching between choices, models often remain anchored to their initial selections (typically one or two specific answer choices) across multiple feedback iterations, even when that choice is demonstrably incorrect. 
We use the same prompts, few-shot demonstrations, answer parsing mechanism, and metrics from \texttt{lm-evaluation-harness} and \texttt{llama-evaluate} where applicable to maintain established evaluation practices and enable fair comparison with prior work. For MMLU, MMLU Pro, PopQA, and TriviaQA, we sample 10\% of the total data points, as running the full dataset through our 10-iteration improvement process would be prohibitively time-consuming. Our preliminary experiments confirmed that this subset yields performance metrics nearly identical to those obtained from the complete dataset. 

\paragraph{Prompt design and experimental controls}
Our experimental framework employs carefully structured prompts to ensure consistent feedback delivery across iterations. For the solver model, we use task-specific system prompts (detailed in \autoref{appendix:prompts}) and construct iterative prompts that include complete interaction history. We investigate whether prompt structure and organization affect feedback incorporation by comparing our standard single-prompt approach against a multi-turn conversation format that structures the same information across multiple dialogue exchanges. In the conversation format, we reformatted the iterative improvement process to mimic natural dialogue, with alternating turns between solver attempts (as user) and feedback responses (as assistant). Results showed marginal differences compared to single-prompt formatting, suggesting that \PHENOMENON{} persists regardless of interaction structure.

Beyond the feedback mechanisms discussed in \S\ref{sec:feedback_mechanisms}, to maintain evaluation integrity while preserving feedback quality, we implement comprehensive answer masking to prevent feedback from directly revealing ground truth solutions. We use ``[masked]'' as the replacement token for filtered content, applying targeted masking that preserves intermediate steps and reasoning guidance while preventing direct answer revelation. For example, we replace numerical answers with ``[masked]'' (e.g., ``The final answer is [masked]'' instead of ``The final answer is 42'') while preserving solution steps  \footnote{We verified that models do not attempt to predict the ``[masked]'' token during inference}.


\paragraph{Models and inference.} 
As for \emph{\underline{solver models,}} we employs strong  models including LLaMA-3.3 70B Instruct \citep{meta_llama3_2024}, Llama-4-Scout-17B-16E, \llamafourbig{} \citep{meta_llama4_2025}, \claude{} and \claude{} with extended thinking \citep{claude3_7}.
Claude 3.7 with extended thinking is a variant that employs an extended reasoning before generating final responses, allowing the model to engage in more deliberate problem-solving through explicit step-by-step thinking.
All models are instruction-tuned versions of their respective base models, specifically optimized for handling natural language instructions and maintaining consistent output formatting.

For the Llama models, during inference, we use temperature 0 to ensure deterministic outputs and conduct inference using vLLM \citep{kwon2023efficientmemorymanagementlarge} with each model's corresponding chat template. All inference is performed on a single H100 instance equipped with eight 80GB GPUs.
For Claude models, we access them through Anthropic's API. For Claude 3.7, we use temperature 0 to ensure deterministic outputs, while for Claude 3.7 with extended thinking, we use temperature 1 as suggested by Claude. We also explore the effects of varying these temperature settings in our experiments to mitigate \PHENOMENON{} (\S\ref{sec:sampling}).

For  \emph{\underline{feedback models,}} we use different models depending on the feedback type. 
For \feedbackthree{}, we utilize GPT-4.1 mini \citep{chatgpt_4.1} as the feedback generation model. From our internal testing, GPT-4.1 mini's feedback performance is on-par with Claude 3.7. Due to the higher cost of Claude 3.7, we use GPT-4.1 mini for generating the strongest feedback. We also considered o4-mini \citep{o3_and_o4} as the feedback generator model due to its reportedly superior reasoning capabilities, but our experiments showed it delivered comparable feedback quality while incurring substantially higher computational costs, leading us to proceed exclusively with GPT-4.1 mini.

\begin{figure*}[t!]
    \centering

    \includegraphics[trim=0.4cm 0cm 0cm 0cm,scale=0.37]{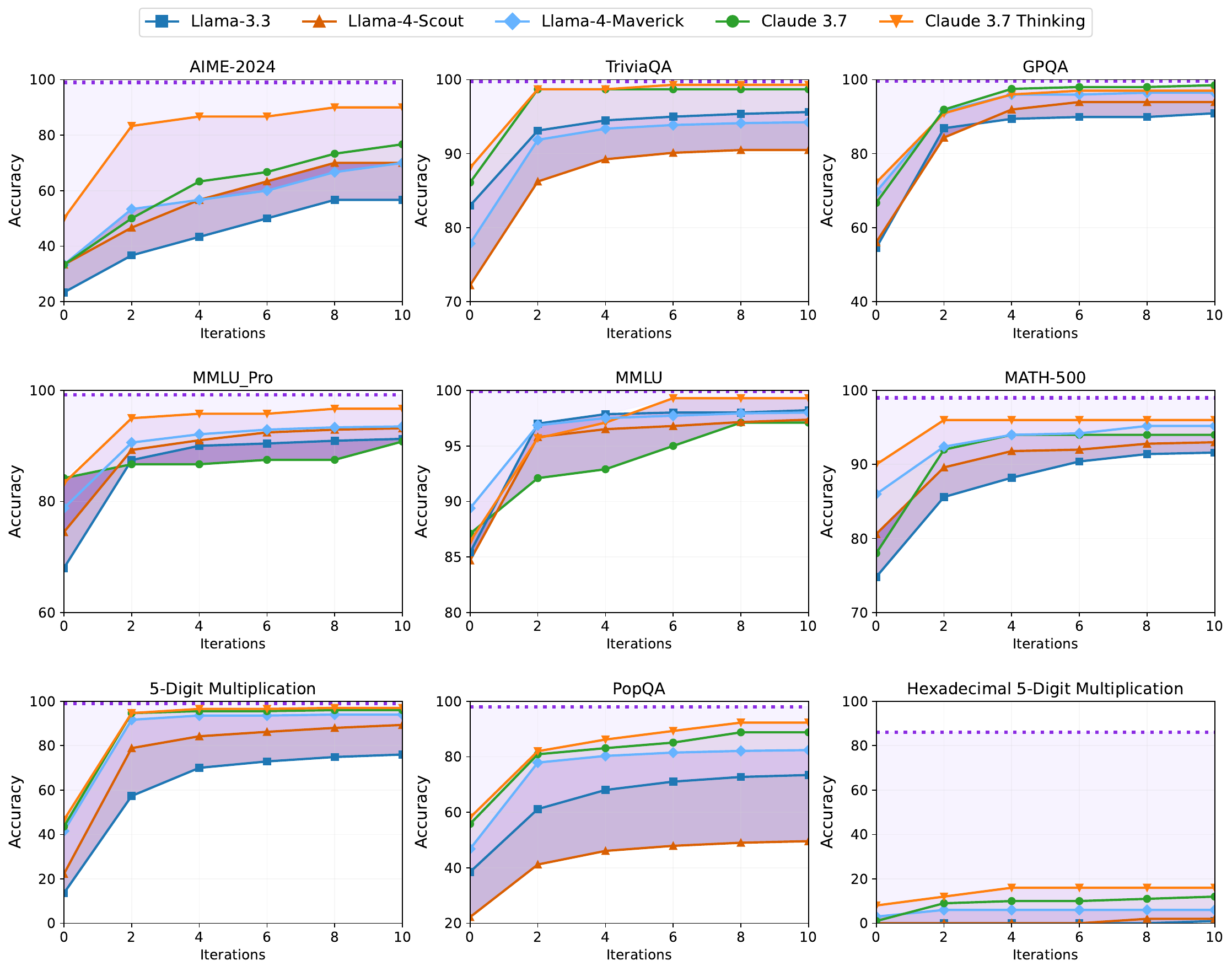}
    \caption{The performance of frontier models we tested with \feedbackthree{} across nine different tasks. Models are given multiple attempts with feedback that incorporates both the final answer and complete solution (when available). The dotted line 
    \protect\tikz[baseline=-0.4ex]{\protect\draw[line width=1.5pt, dash pattern=on 1pt off 1pt, mypurple] (0,0) -- (0.33cm,0);}
    represents the target accuracy that models could theoretically achieve if they fully incorporated all feedback (details in \S\ref{sec:categorization}). \textbf{Results demonstrate that despite strong feedback, models consistently plateau below their target accuracy across all tasks.}}
    \label{fig:results}
\end{figure*}

\subsection{Main findings} \label{sec:results}
\paragraph{\PHENOMENON{} persists across model scales and tasks.}
\autoref{fig:results} shows results using our strongest feedback mechanism (\feedbackthree{}) across all datasets and all models. 
Our results reveal a striking pattern: 
Despite receiving high-quality feedback, all solver models consistently plateau below their target accuracy, which is the accuracy of an ideal model if it successfully incorporates all the given feedback (see \S\ref{sec:categorization} for calculation details). 
Claude 3.7 Thinking achieves the highest initial accuracy on several tasks (AIME, TriviaQA, GPQA, and MATH-500), while Claude 3.7 shows competitive performance across most benchmarks. However, both Claude variants exhibit the same fundamental plateauing behavior as the Llama models.
The performance typically improves rapidly through the first 2-4 iterations before significantly slowing down. This \PHENOMENON{} is particularly pronounced on complex reasoning tasks like AIME and GPQA, where even the best-performing models remain 15-25\% and 3-8\% below their respective theoretical ceilings despite 10 correction opportunities.
The synthetic tasks reveal particularly interesting patterns across model families. In the standard 5-Digit Multiplication task, both Claude models reach near-perfect accuracy after significant initial improvement, outperforming the Llama models. However, the Hexadecimal-5-Digit-Multiplication task reveals extreme difficulty with feedback incorporation across all models---no model exceeds 20\% accuracy even after 10 iterations, highlighting severe limitations in feedback integration for counterfactual arithmetic systems. \footnote{While it is true that models' imperfect understanding of hexadecimal arithmetic leads to imperfect feedback quality, this is reflected in the lower theoretical ceiling shown in the figure. Even accounting for this limitation, models still fall short of what they could achieve if they fully incorporated the available feedback.}

\begin{figure}[h!]
    \centering
\includegraphics[trim=0.8cm 0.9cm 0cm 0cm, scale=0.15]{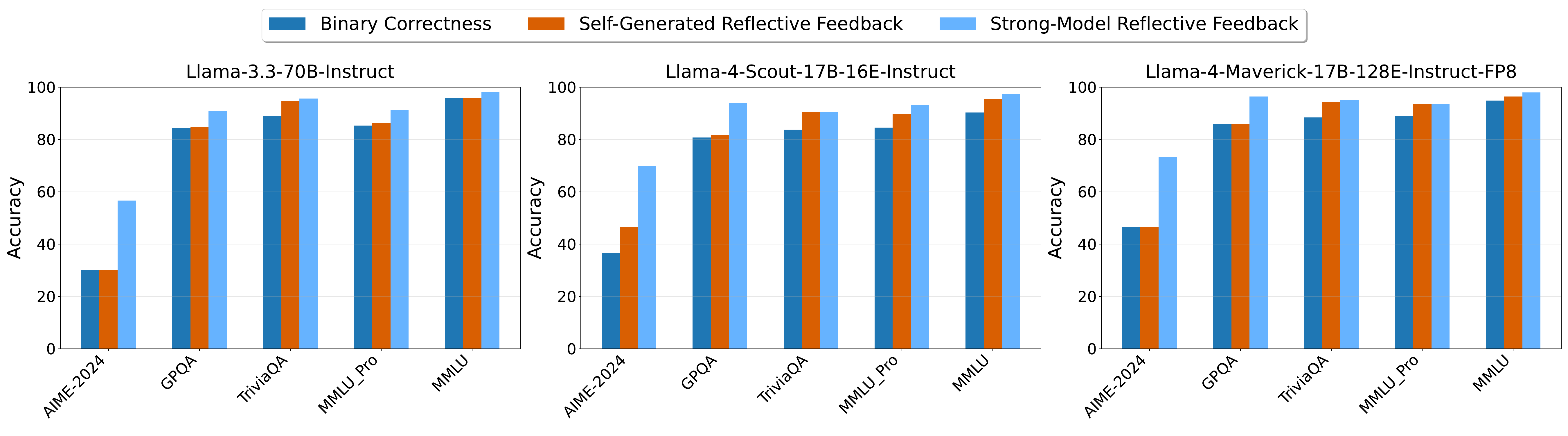}
    \caption{Performance comparison across benchmark datasets using different feedback mechanisms with \llamathreeshort{}, \llamafoursmallshort{} and \llamafourbigshort. \textbf{Model performance progressively improves as feedback quality increases from \feedbackone{} to \feedbackthree{}.}}
    \label{fig:feedback_type}
\end{figure}

\textbf{Feedback quality significantly impacts performance gains.}
\autoref{fig:feedback_type} illustrates how different feedback mechanisms affect performance across models and tasks. Due to the cost of running extensive experiments with Claude models, we focus the feedback quality analysis on the Llama model families. All tasks show clear benefits from increasingly sophisticated feedback.

The impact of high-quality feedback is most pronounced on complex reasoning tasks. For AIME, MMLU Pro, and GPQA, \feedbackthree{} outperforms binary feedback by significant margins across all models. \llamafourbigshort{} shows the strongest overall performance, achieving 73.3\% accuracy on AIME and 96.5\% on GPQA with \feedbackthree----improvements of +26.7\% and +10.6\% over binary feedback, respectively. \llamafoursmallshort{} demonstrates the largest relative gains, particularly on AIME (+33.3\%) and GPQA (+13.1\%), compared to \llamathreeshort{} which shows more modest improvements (+26.7\% on AIME, +6.6\% on GPQA). Nevertheless, despite these substantial improvements, all models still plateau significantly below their theoretical performance ceiling.

\section{Analysis of \PHENOMENON{}}\label{sec:analysis}
We conduct a deeper analysis to better understand \PHENOMENON{}. We first categorize different cases where models fail to correct their mistakes despite multiple rounds of feedback (\S\ref{sec:categorization}), then examine the extent to which we can alleviate this problem with sampling strategies (\S\ref{sec:sampling}), and finally, we present several hypotheses and the experiments to understand \PHENOMENON{} (\S\ref{sec:understanding}).


\subsection{Feedback integration failures dominate persistent self-improvement errors}\label{sec:categorization}

\paragraph{Error category development.} We manually examine cases where LLMs fail to improve despite receiving high-quality feedback and identifies three main categories of error: (1) Most critically for our analysis, \textbf{feedback resistance failures} represent cases where models fail to accurately incorporate feedback despite multiple iterations. (2) \textbf{Feedback quality issues} encompass cases where the provided feedback is incorrect, ambiguous, or fails to address the key problematic steps in the solution. This can still occur because the generated feedback might miss crucial conceptual errors or introduce new inaccuracies, even though ground-truth is provided to the feedback generator. (3) We maintain an \textbf{``Other''} category for cases that don't clearly fit into either of the above categories. From our initial examination, this includes cases where the problem itself contains ambiguities or the solution is conceptually correct but fails due to style or formalization issues (e.g., providing the correct information but not in the expected format required by the evaluation metric).

\begin{wraptable}[14]{r}{0.60\textwidth}
\small
\vspace{-0.45cm}
\caption{Distribution of error categories (\%) of unsolved problems after 10 iterations of self-improvement classified by o4-mini. FR: Feedback Resistance, FQ: Feedback Quality, OTH: other issues. 
}\label{tab:error_categories}
\begin{tabular}{l|l|ccc}
\toprule
Dataset & Solver Model & FR & FQ & OTH \\
\midrule
MMLU Pro & Claude 3.7 & 64.6 & 28.0 & 7.4 \\
 & Claude 3.7 Thinking & 62.8 & 30.8 & 6.4 \\
GPQA & Claude 3.7 & 100.0 & 0.0 & 0.0 \\
 & Claude 3.7 Thinking & 85.7 & 14.3 & 0.0 \\
TriviaQA & Claude 3.7 & 72.4 & 25.0 & 2.6 \\
 & Claude 3.7 Thinking & 71.7 & 28.3 & 0.0 \\
AIME 2024 & Claude 3.7 & 100.0 & 0.0 & 0.0 \\
 & Claude 3.7 Thinking & 100.0 & 0.0 & 0.0 \\
\bottomrule
\end{tabular}
\end{wraptable}

\paragraph{Automated error categorization and validation.} To systematically categorize errors at scale, we used OpenAI o4-mini as an automated annotator to classify complete self-improvement trajectories from \claudeshort{} and \claudethinkshort{} according to these predefined categories. To validate this automated approach, we conducted rigorous manual verification by randomly sampling 50 errors from each task and having two human annotators independently label them according to our defined categories. Our verification showed 96\% agreement between human annotators and o4-mini's classifications, significantly higher than the 78\% agreement rate achieved with GPT-4.1 mini on the same samples. This confirms o4-mini's reliability for this analysis task.

\paragraph{Feedback resistance dominates error patterns.} \autoref{tab:error_categories} presents the distribution of error categories across different tasks, as classified by o4-mini. Feedback resistance is consistently the dominant category across all tasks, accounting for 62.8-100\% of errors. This finding suggests that the core challenge in achieving perfect performance lies not in the quality of feedback or problem complexity, but in fundamental limitations of how models process and incorporate corrective feedback. Detailed examples of each error category are provided in \autoref{appendix:error_categorization}.

\subsection{Mitigating \PHENOMENON{} with sampling strategies}\label{sec:sampling}
Given the persistent plateau in performance we observed across models and tasks, a natural question arises: \textit{can we mitigate \PHENOMENON{} through existing strategies?} We explore sampling techniques as a potential solution to help models overcome their apparent resistance to feedback.

\paragraph{Progressive temperature increases show limited effectiveness.} We first explore temperature-based sampling strategies where the sampling temperature increases with the iterations: 0.0 for iteration 0, 0.15 for iteration 1, 0.3 for iteration 2, and so forth. While other schedules (e.g., exponential increase, fixed higher temperature) could be explored, we chose this linear progression as a simple baseline that gradually introduces diversity while preserving early deterministic behavior. We hypothesize that progressive temperature increases would help models generate more diverse outputs, allowing them to escape from local optima in their output distributions and become more receptive to feedback. 

\begin{figure}[ht]
\centering
\includegraphics[trim=0.8cm 0.0cm 0cm 0cm, scale=0.35]{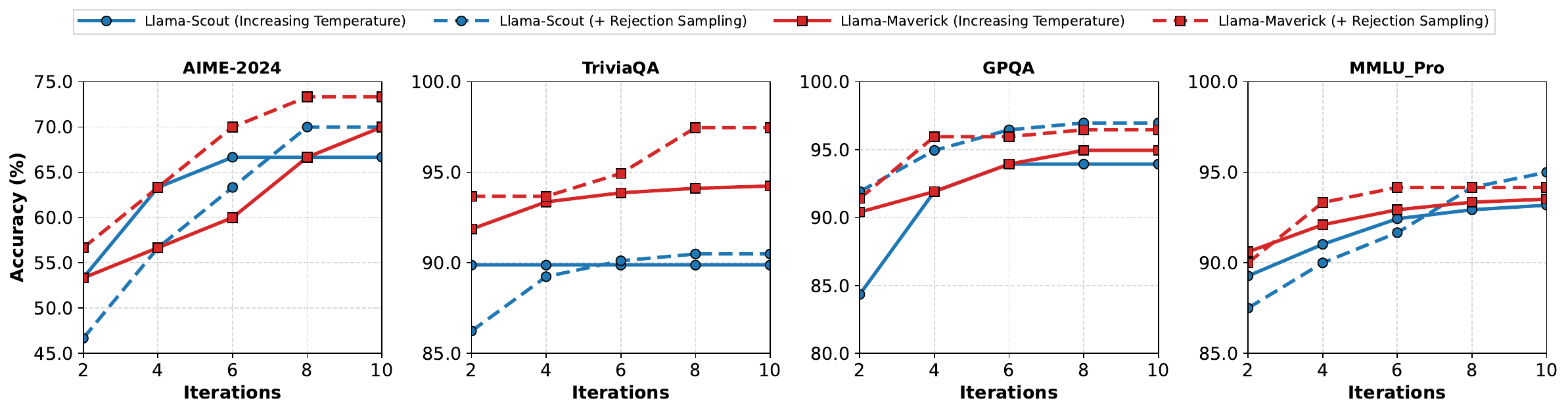}
\caption{Results of using progressively increasing temperature and rejection sampling with \llamafoursmallshort{} and \llamafourbigshort{}. \textbf{Rejection sampling can provide additional improvements over temperature-based sampling alone across both multiple-choice and non-multiple-choice tasks.}}
\label{fig:sampling}
\end{figure}

Due to the cost of running extensive experiments with Claude models, we focus the feedback quality analysis on \llamafoursmallshort{} and \llamafourbigshort{}. As shown in \autoref{fig:sampling}, this approach alone produced minimal improvements compared to the baseline in \autoref{fig:results}. Analysis of logs revealed that while increased temperature successfully diversified model outputs, the additional exploration often failed to converge on correct answers due to the vast search space of possible responses.

\paragraph{Combining temperature increases with rejection sampling yields better results.} To enhance performance further, we implement a more targeted approach that combines increased temperature with rejection sampling. This method explicitly forces the model to explore new solution paths while avoiding previous attempts. Specifically, we instruct the model to generate 25 answers, and remove final answers that occurred in previous iterations (which by construction is incorrect, since we only continue iterating on problems that remain unsolved). If no answer remains after this filtering process, we randomly select one from those 25 answers. Otherwise, we randomly select one of the remaining novel answers as the final prediction.

As shown in \autoref{fig:sampling}, the combined strategy yields substantive performance gains across both multiple-choice datasets and non-multiple-choice datasets compared to the baseline that only increases the temperature. While the magnitude of these improvements varied, the consistent pattern suggests that forcing models to explore new solution paths by rejecting previously used answers is beneficial. Despite these gains, we observed that all datasets still fall short of the target accuracy, indicating that sampling strategies alone cannot fully resolve model resistance to feedback.

\subsection{Understanding \PHENOMENON{}} \label{sec:understanding}
Our sampling strategies (\S\ref{sec:sampling}), though promising, did not eliminate \PHENOMENON{} entirely. Developing more effective interventions requires a deeper understanding of the fundamental causes. In this section, we investigate several hypotheses for why models resist incorporating feedback despite multiple correction opportunities. Throughout our analyses, we provide error bars. They represent the standard error of a binomial proportion, calculated as $\sqrt{\text{acc} \cdot (1-\text{acc}) / n}$, where $n$ is the number of samples in each evaluated group.

\paragraph{Model confidence and \PHENOMENON{}}
Could excessive model confidence explain resistance to feedback in \PHENOMENON{}?
To test this, we measure confidence using \emph{semantic entropy}~\cite{semantic_entropy}, a method that captures uncertainty at the meaning level rather than surface form variations. Unlike token-level probability measures, semantic entropy accounts for the fact that models can express the same meaning in multiple ways, providing a more robust measure of true uncertainty.

To compute semantic entropy for each model response, we first generate multiple outputs $\{s_1, s_2, \ldots, s_n\}$ by sampling the model $n=50$ times with temperature $0.7$, where each $s_i$ represents a complete answer given the feedback. We then cluster these outputs using final answers, which we find to have similar results as semantic equivalence grouping using bidirectional entailment. For each semantic cluster $C_i$, we estimate its probability as $\hat{P}(C_i|x) = \sum_{s \in C_i} P(s|x) / \sum_{j=1}^K \sum_{s \in C_j} P(s|x)$, where $x$ represents the input (question, previous answer, and feedback), and compute semantic entropy as $H = -\sum_i \hat{P}(C_i|x) \log \hat{P}(C_i|x)$. Lower entropy indicates the model consistently generates semantically equivalent answers (high confidence), while higher entropy suggests diverse semantic meanings (low confidence).

We conduct this analysis across 5 digits
multiplication, GPQA, MATH, MMLU-Pro, and TriviaQA using semantic entropy with bucket size 0.2. \autoref{fig:semantic_entropy} shows some of the results (full results in \autoref{appendix:semantic_entropy}) and reveals a consistent pattern: absolute improvement rate (the green trends) increases with semantic entropy, rising from near-zero at low entropy (high confidence) to 0.4-0.8 at higher entropy levels (low confidence). 
These results suggest that models with lower confidence (higher semantic entropy) have both more room for improvement and greater receptiveness to feedback, while highly confident models show minimal improvement despite receiving the same quality feedback.

\begin{figure*}[t]
\centering
\includegraphics[width=\textwidth]{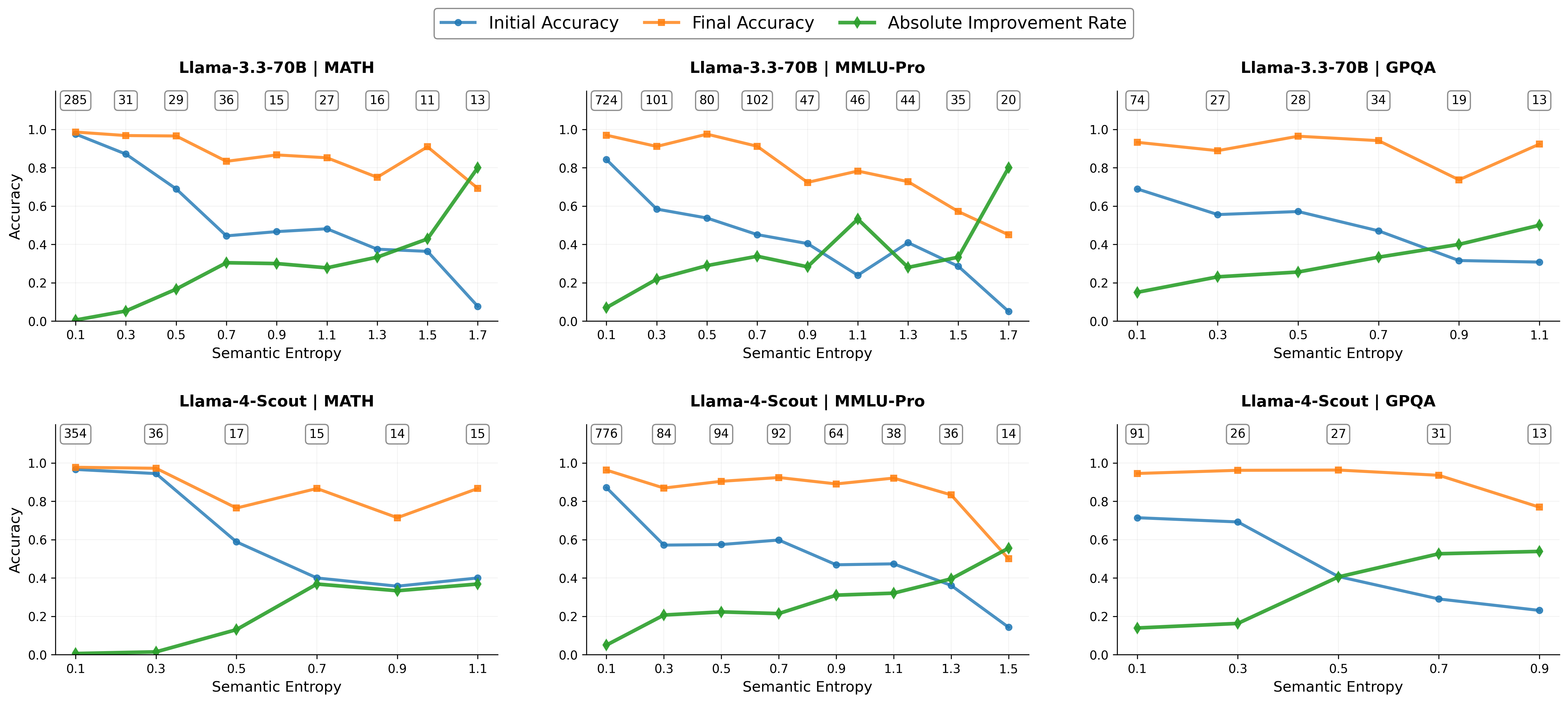}
\caption{
Relationship between semantic entropy and feedback incorporation across three benchmark tasks (MATH, MMLU-Pro, GPQA) for Llama-3.3-70B and Llama-4-Scout models. Semantic entropy (x-axis) measures model uncertainty, with lower values indicating higher confidence. Three metrics are shown: \textcolor{blue}{initial accuracy} before any feedback (blue), \textcolor{orange}{final accuracy} after iterative feedback (orange), and \textcolor{green!60!black}{absolute improvement rate} calculated as $(Final - Initial) / (1 - Initial)$ (green). Numbers in boxes indicate sample sizes per bucket. The \textcolor{green!60!black}{absolute improvement rate} generally increases with semantic entropy across all tasks and models, indicating that models with lower initial confidence (higher entropy) show greater relative improvements from feedback.}
\label{fig:semantic_entropy}
\end{figure*}

\textbf{Model self-perception versus actual behavior.}
To probe deeper into \PHENOMENON{} mechanisms, we directly asked solver models about their understanding of feedback and willingness to incorporate it. After receiving feedback, we prompt models with: (1)~``Do you understand this feedback?'' and (2)~``Will you update your belief or understanding about this problem?''

Surprisingly, across all models and tasks, solver models consistently claimed to understand the feedback ($>95\%$ ``yes'' responses) and expressed willingness to update their beliefs ($>96\%$ ``yes'' responses). However, in subsequent iterations, these same models failed to actually incorporate the feedback. This disconnect between stated intention and actual behavior reveals a fundamental issue: models exhibit self-assessment failure where they believe they are incorporating feedback while demonstrably failing to do so (the prompts used for probing and example conversations can be found in \autoref{appendix:example_model_self_perception}). This disconnect between stated intention and actual behavior reveals a fundamental implementation failure: models express both understanding and willingness to change but fail to execute these changes in practice.

\begin{wrapfigure}[20]{l}{0.4\textwidth}
    \centering
    \begin{tabular}{c}
        \includegraphics[trim=1.5cm 0.0cm 0.8cm 2.0cm, scale=0.35]{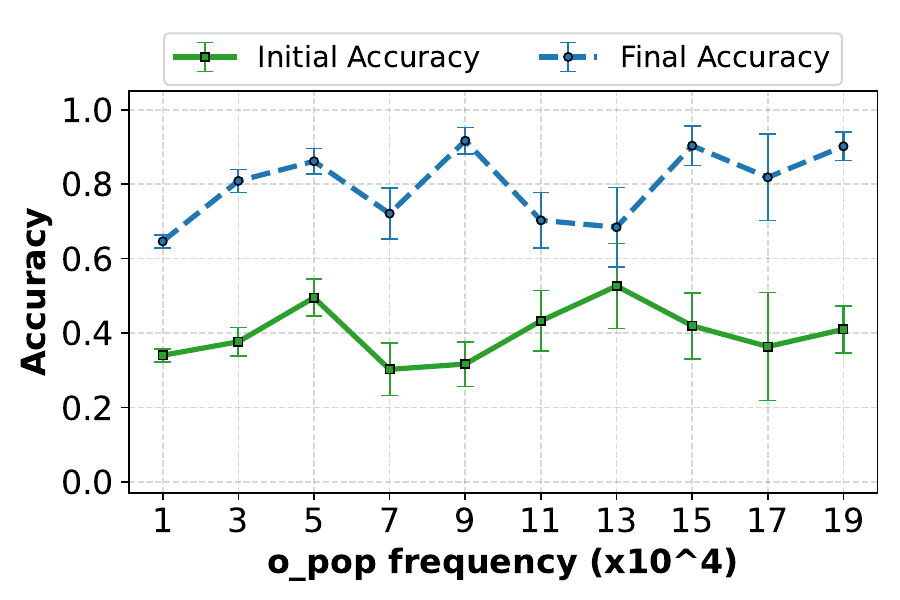} \\
        \includegraphics[trim=1.5cm 0.8cm 0.8cm 0.5cm, scale=0.35]{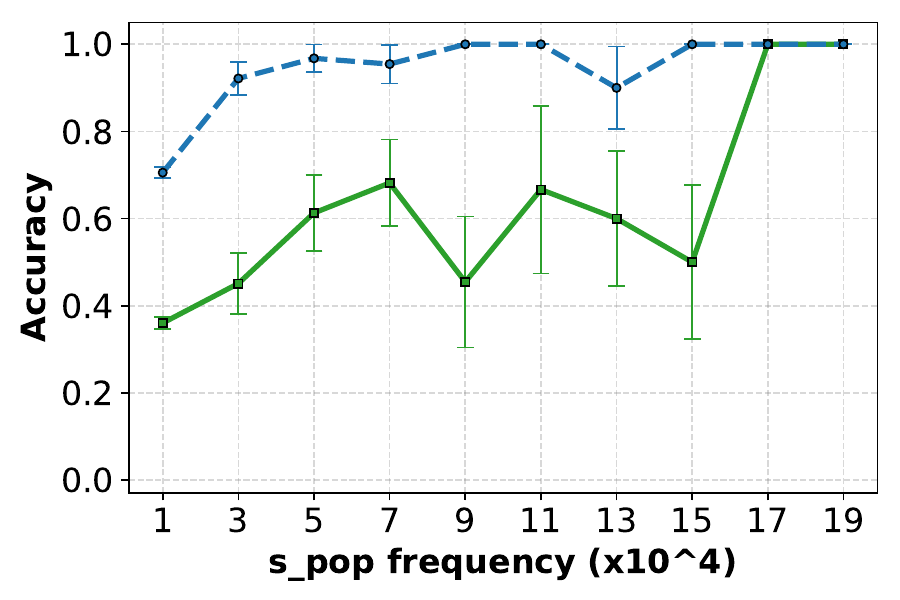} \\
    \end{tabular}
    \caption{Accuracy of Llama3.3 on PopQA with respect to s\_pop and o\_pop}
    \label{fig:familiarty}
\end{wrapfigure}

\paragraph{Data familiarity and \PHENOMENON{}}
Prior work \citep{ember, mallen2022trustlmretrieval} suggests that language models perform better with familiar entities and topics encountered frequently during training. Are these models more resistant to feedback about familiar entities? We investigated whether this familiarity bias contributes to \PHENOMENON{} using PopQA's popularity metrics as proxies for entity familiarity. 
This dataset includes monthly Wikipedia page views for subject entities (s\_prop) and object entities (o\_prop). We then analyzed how accuracy changes during iterative self-improvement correlate with the popularity of the subject (s\_prop) and object (o\_prop) entities.


We use \llamathreeshort{} for this experiment since it was released closer to the publication date of PopQA. These popularity metrics provide a particularly relevant measure of potential training data frequency for this model. As shown in \autoref{fig:familiarty}, we found no consistent pattern between entity popularity and accuracy. We provide further supporting evidence in \autoref{appendix:bias} with additional statistical testing and alternative popularity metrics.

\paragraph{Further analyses} 
We also explored whether problem complexity, as measured by the number of expected reasoning steps, correlates with \PHENOMENON{} since prior works \citep{dziri2023faithfatelimitstransformers} show that longer reasoning trajectories may impede problem-solving in LLMs. Additionally, we investigated whether certain questions consistently induced reluctance to feedback across different models because of their inherent characteristics. For both hypotheses, our controlled experiments yielded negative results, with minimal correlation between reasoning complexity and stubbornness, and little overlap in stubborn questions across models. Complete analyses are provided in \autoref{appendix:reasoning_complexity} and \autoref{appendix:model_type}.

\section{Related Work}
\paragraph{Self-Improvement with LLMs.} 
Self-improvement in artificial intelligence has evolved significantly over time, with roots predating the current LLM era. Early work explored using Generative Adversarial Networks (GANs) to enable NLP systems to improve through self-generated feedback \citep{subramanian2017adversarial, yu2017seqgan}.
The emergence of LLMs has dramatically expanded both the scope and capabilities of self-improvement techniques. Modern applications span diverse domains including code generation \citep{zelikman2024selftaughtoptimizerstoprecursively}, reasoning tasks \citep{self-refine, nathani2023mafmultiaspectfeedbackimproving}, instruction following \citep{wang2023selfinstruct, selfee2023}, and many others \citep{chen2024iterativetranslationrefinementlarge, pryzant2023automaticpromptoptimizationgradient, Yin_2024}.
Approaches to achieve self-improvement vary in their focus - some concentrate on training time improvements, where models learn to self-improve \citep{kumar2024traininglanguagemodelsselfcorrect} or generate additional training data \citep{self-play, self-rewarding}, while others emphasize inference time improvements, often incorporating feedback mechanisms \citep{self-refine, Reflexion} though sometimes utilizing other models without explicit feedback \citep{welleck2022selfcorrect}.
While several studies cast doubt on whether off-the-shelf LLMs possess the ability for intrinsic self-improvement (i.e., the ability to self-improve without using any external ground-truth feedback) \citep{cant, jiang2024selfincorrectllmsstrugglediscriminating}, there is consensus that LLMs can self-improve when such feedback is available \citep{can_correct, pan2023automaticallycorrectinglargelanguage}.
In this work, we probe the limits of self-improvement with external ground-truth feedback and investigate what prevents LLMs from fully integrating feedback.


\paragraph{The elasticity–plasticity dilemma}
LLMs often face an elasticity–plasticity dilemma---a trade-off between retaining prior knowledge and integrating new information across various adaptation scenarios. In continual learning, LLMs exhibit catastrophic forgetting, as new knowledge can overwrite or interfere with old knowledge, especially when updates conflict with the model’s existing beliefs \citep{clemente2025stubbornness, luo2023empirical, ming2025faithevallanguagemodelstay}. 
Similarly, in model editing, a handful of targeted edits can successfully inject new facts, but beyond only a few edits the model’s performance on unrelated queries and general benchmarks deteriorates, indicating that extensive edits irreversibly distort the model’s broader knowledge network \citep{li2024should, mitchell2022memory}. Finally, alignment fine-tuning techniques such as instruction tuning and reinforcement learning from human feedback (RLHF) encounter a similar dilemma: while they instill desired behaviors, the tuned models often remain elastically tied to their pre-trained behavior distribution or can be coerced (via adversarial prompts) to revert to undesirable outputs, suggesting that alignment can be brittle or superficial \citep{ji2024resist, zou2023universal}. In this paper, we revisit this dilemma through the lens of feedback integration during self-improvement. We also try to unveil the factors that control this dilemma through analysis. 

\paragraph{Feedback for LLMs.}
Prior work has explored various approaches for providing feedback to LLMs during self-improvement processes. Feedback mechanisms can be broadly categorized into intrinsic feedback, where LLMs evaluate their own outputs through prompting \citep{madaan2023self, dhuliawala2024chain, wu2024largelanguagemodelsselfcorrect, varshney2023stitchtimesavesnine}, and extrinsic feedback leveraging additional tools \citep{gou2024criticlargelanguagemodels, jiang2023selfevolvecodeevolutionframework, chen2023teachinglargelanguagemodels}, information sources \citep{zhao-etal-2023-verify, yu2023improvinglanguagemodelsplugandplay}, or even ground-truth answer \citep{kim2023languagemodelssolvecomputer, Reflexion}.
While research demonstrates that generating correct feedback is the key for LLM self-improvement \citep{kamoi2024llmsactuallycorrectmistakes, can_correct}, challenges remain in how effectively LLMs incorporate this feedback. Studies suggest that LLMs may struggle to accept new information that contradicts their prior knowledge \citep{wu2024clashevalquantifyingtugofwarllms}, handle refuting instructions \citep{yan2024refutebenchevaluatingrefutinginstructionfollowing}, or may be led astray by misleading feedback \citep{xu2024earthflatbecauseinvestigating, wang2023chatgptdefendbelieftruth}. Different from previous work, we focus specifically on high-quality and even perfect feedback, investigating why LLMs fail to fully incorporate such helpful feedback.

\section{Limitations}
\paragraph{Better understanding of \PHENOMENON{}}
While our study identifies key insights into \PHENOMENON{}—including the correlation between semantic entropy and feedback resistance and the disconnect between models' stated understanding and actual behavior, we lack a definitive mechanistic explanation for why models resist incorporating feedback. A better understanding of \PHENOMENON{} likely involves complex interactions between feedback understanding, instruction following, and belief updating. Future work would benefit from more sophisticated mechanistic interpretability techniques, such as causal intervention methods and circuit analysis \citep{yang2025llmsadmitmistakesunderstanding, meng2023locatingeditingfactualassociations}, to understand the specific computational pathways through which feedback resistance emerges and persists across model architectures.

\paragraph{Limited mitigation strategies}
Despite extensive experimentation with sampling strategies, our attempts to fully mitigate feedback friction yields only modest improvements. The most promising next step appears to be supervised fine-tuning or reinforcement learning approaches that could (1) enhance the solver model's receptiveness to feedback incorporation, and (2) improve the feedback generator's ability to provide more effective guidance \citep{yao2024retroformerretrospectivelargelanguage}. However, the computational constraints of our experimental setup, particularly the large model sizes we tested (including 70B+ parameter models like Claude 3.7 and Llama-4-Maverick) and the associated computational costs, prevented us from conducting fine-tuning experiments that might meaningfully address feedback resistance.

\section{Conclusion}
Our study reveals a fundamental limitation in LLMs' ability to incorporate external feedback. Despite receiving high-quality feedback over multiple iterations, models consistently plateau below their theoretical performance ceiling across diverse reasoning tasks. We investigated a few sampling strategies that mitigated this issue, but did not eliminate this stubborn plateau.
Despite extensive analysis, the precise mechanisms underlying feedback resistance remain elusive, which we leave to future work. Understanding and addressing these limitations remain essential for developing more adaptable AI systems capable of genuine and sustained self-improvement.

\section*{Acknowledgment}
This work is supported by ONR grant (N00014-241-2089). 
The GPUs were provided by the DSAI and ARCH clusters. 
We sincerely thank Yuqing Yang, Zhengping Jiang, and the broader JHU community for discussions and feedback.





\bibliography{ref}


\appendix

\section{Prompts and evaluation details for problem solving and feedback generation} \label{appendix:prompts}
\subsection{Prompts used for the solver model}
We developed different prompting strategies for the solver model to incorporate feedback and generate new solutions across iterations. The system prompts in \autoref{fig:sys_prompt} were used for the solver model across different tasks: 

\begin{figure}[h!] \centering \includegraphics[width=1\linewidth]{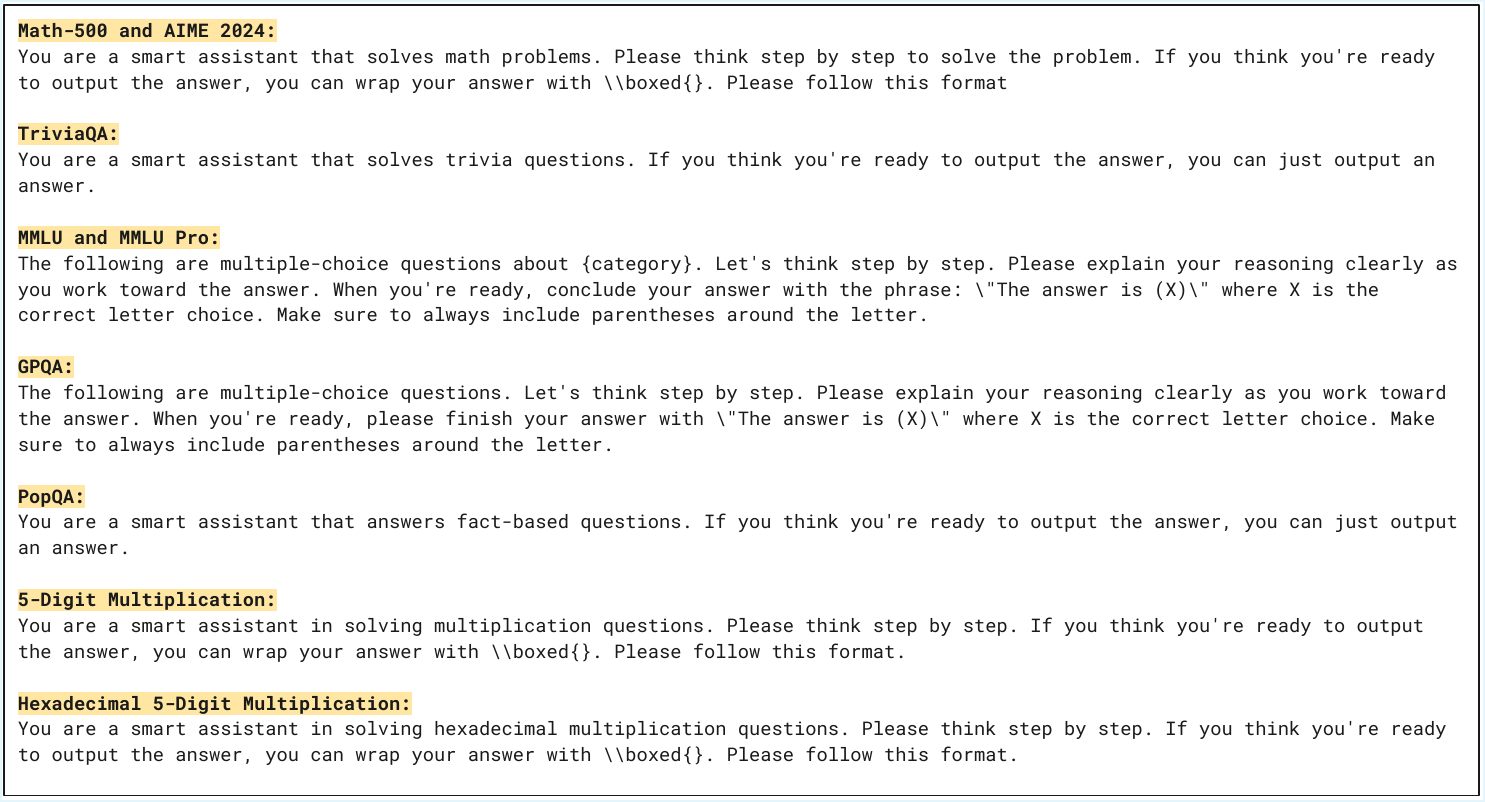} \caption{System prompts used for the solver model across all tasks.} \label{fig:sys_prompt} \end{figure} 

\begin{figure}[h!]
    \centering
    \includegraphics[width=1\linewidth]{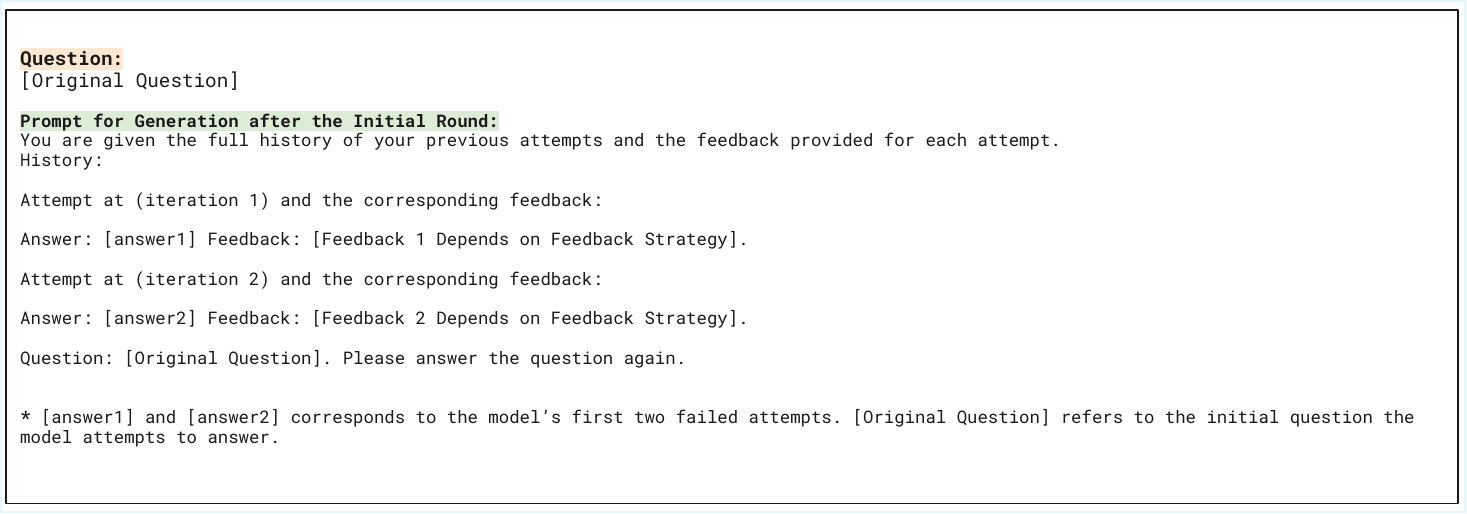}
    \caption{Prompt used for iterative self-improvement}
    \label{fig:prompt_generate}
\end{figure}

At the initial generation round, we directly use the question as the prompt for the solver model. For multiple-choice questions, we format the question by concatenating the question and answer choices with their corresponding labels (i.e., A to D for MMLU and GPQA, A to J for MMLU Pro). In subsequent rounds, we provide the solver model with its complete previous history and the corresponding feedback from the feedback generator model. We clearly label each iteration so the solver model can track all its previous attempts. The general template for the iterative prompt structure is provided in \autoref{fig:prompt_generate}. 

\subsection{Evaluation details for problem solving}

We employ few-shot prompting across all tasks to provide consistent context for the solver model. For TriviaQA and PopQA, we randomly sample 5 questions without replacement as few-shot examples. For MMLU and MMLU Pro, we similarly sample 5 questions from the corresponding question category to ensure domain-relevant examples.

For PopQA, we employ an LLM-as-a-judge approach \cite{llm-as-a-judge} to assess answer correctness. This is necessary because PopQA provides limited answer aliases (extensive alternative phrasings for exact string matching) compared to TriviaQA. Without this approach, models would be penalized for minor formatting differences rather than genuine comprehension errors, leading to an underestimation of their true problem-solving capabilities. For other tasks, we follow the same evaluation metrics provided by lm-eval-harness.

\subsection{Prompts used for feedback generation}

We implement three distinct feedback generation strategies as described in \S\ref{sec:feedback_mechanisms}. For \feedbackone{}, we provide minimal information: ``Your answer was incorrect. Please answer the question again.''

\begin{figure}[h!]
    \centering
    \includegraphics[width=1\linewidth]{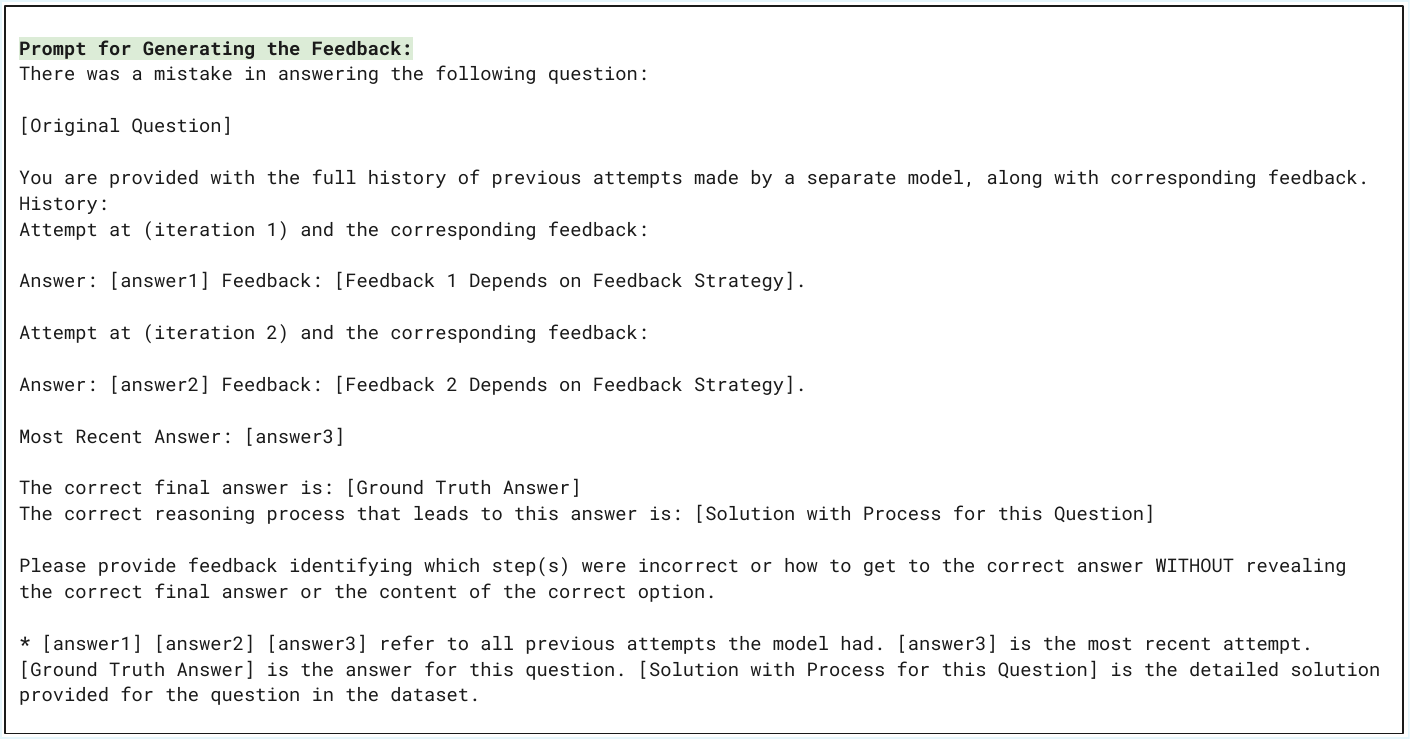}
    \caption{Prompt used for generating the feedback.}
    \label{fig:feedback_prompt}
\end{figure}

For \feedbacktwo{} and \feedbackthree{}, we employ identical prompt templates that differ only in the model used for generation. The feedback generator receives the complete interaction history, including all previous solver attempts and corresponding feedback. When available, we provide the feedback model with detailed solution explanations that justify the correct answer; for datasets lacking such explanations, we provide only the ground truth answer. This approach ensures the feedback model has sufficient context to generate targeted, informative guidance while maintaining consistency across feedback types, and its template is shown in \autoref{fig:feedback_prompt}.

\subsection{Answer masking in feedback}

To ensure fair evaluation, we implement comprehensive answer masking to prevent feedback from directly revealing ground truth solutions while preserving feedback quality. Our approach allows feedback to contain detailed solution steps and guidance but strictly prohibits explicit disclosure of final answers. We use ``[masked]'' as the replacement token for filtered content.

\paragraph{Multiple-choice questions.} We mask all possible representations of the correct choice letter. For example, if the correct answer is A, we filter variants including (A), \texttt{\textbackslash boxed\{A\}}, **A**, etc.

\paragraph{Open-ended questions.} 
For TriviaQA, we filter all terms matching the words in ``aliases'' and ``normalized aliases'' answer fields. For PopQA, we mask entries from the ``possible answers'' answer field. For mathematical tasks (5-digit multiplication and MATH-500), we mask standalone numerical answers and those in \texttt{\textbackslash boxed\{\}} notation. Hexadecimal multiplication follows similar patterns. For multiplication tasks, we additionally mask intermediate partial products to prevent reduction to simple addition problems (detailed in \autoref{appendix:synthetic_task}).

\subsection{Error Categorization Prompt}
\begin{figure}[h!]
    \centering
    \includegraphics[width=1\linewidth]{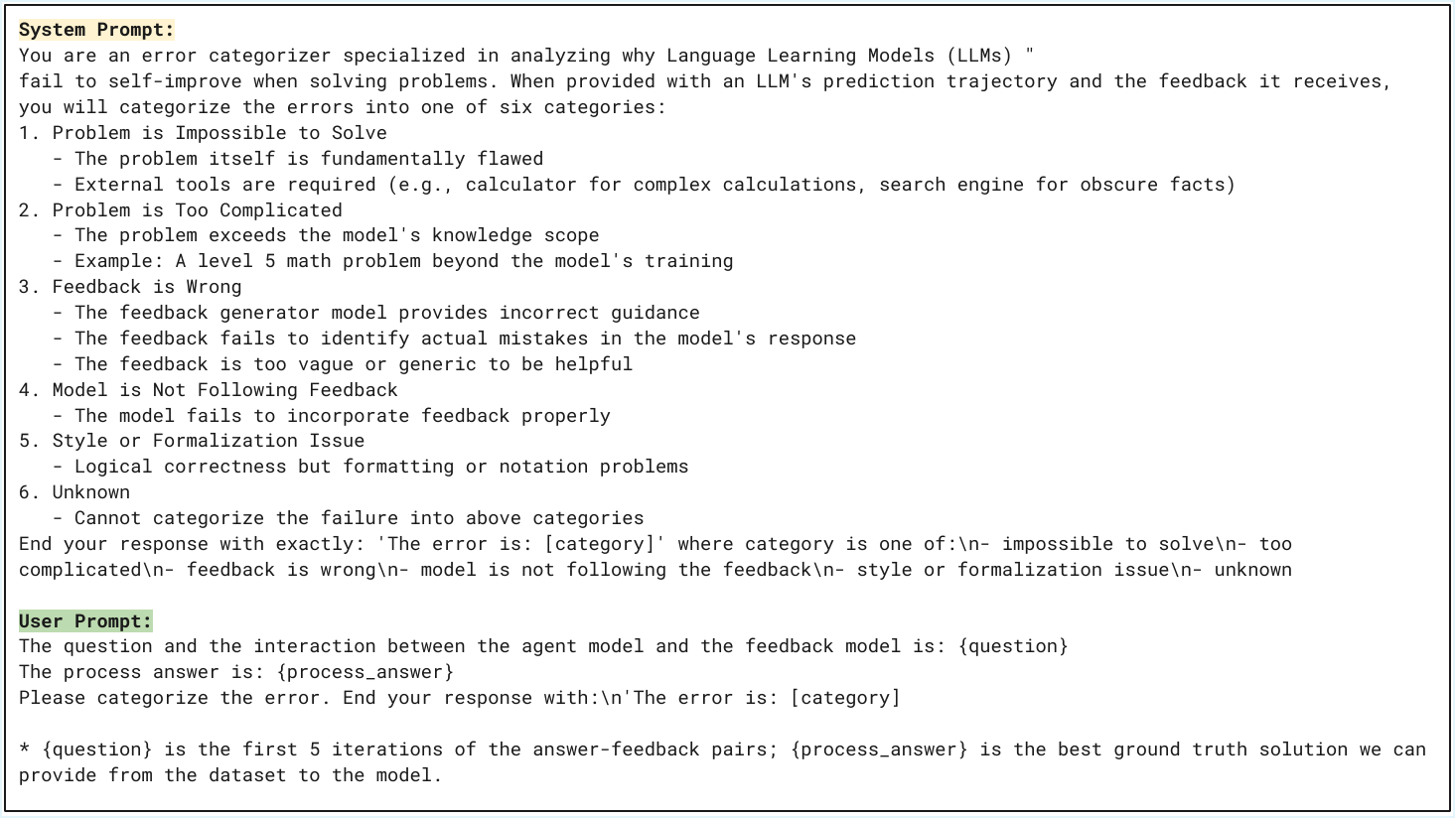}
    \caption{Error Categorization}
    \label{fig:cat}
\end{figure}

The prompt template used for categorizing persistent model errors after 9 iterations is shown in \autoref{fig:cat}.

\section{Error categorization examples from iterative self-improvement} \label{appendix:error_categorization}

\begin{figure}[h!]
    \centering
    \includegraphics[width=1\linewidth]{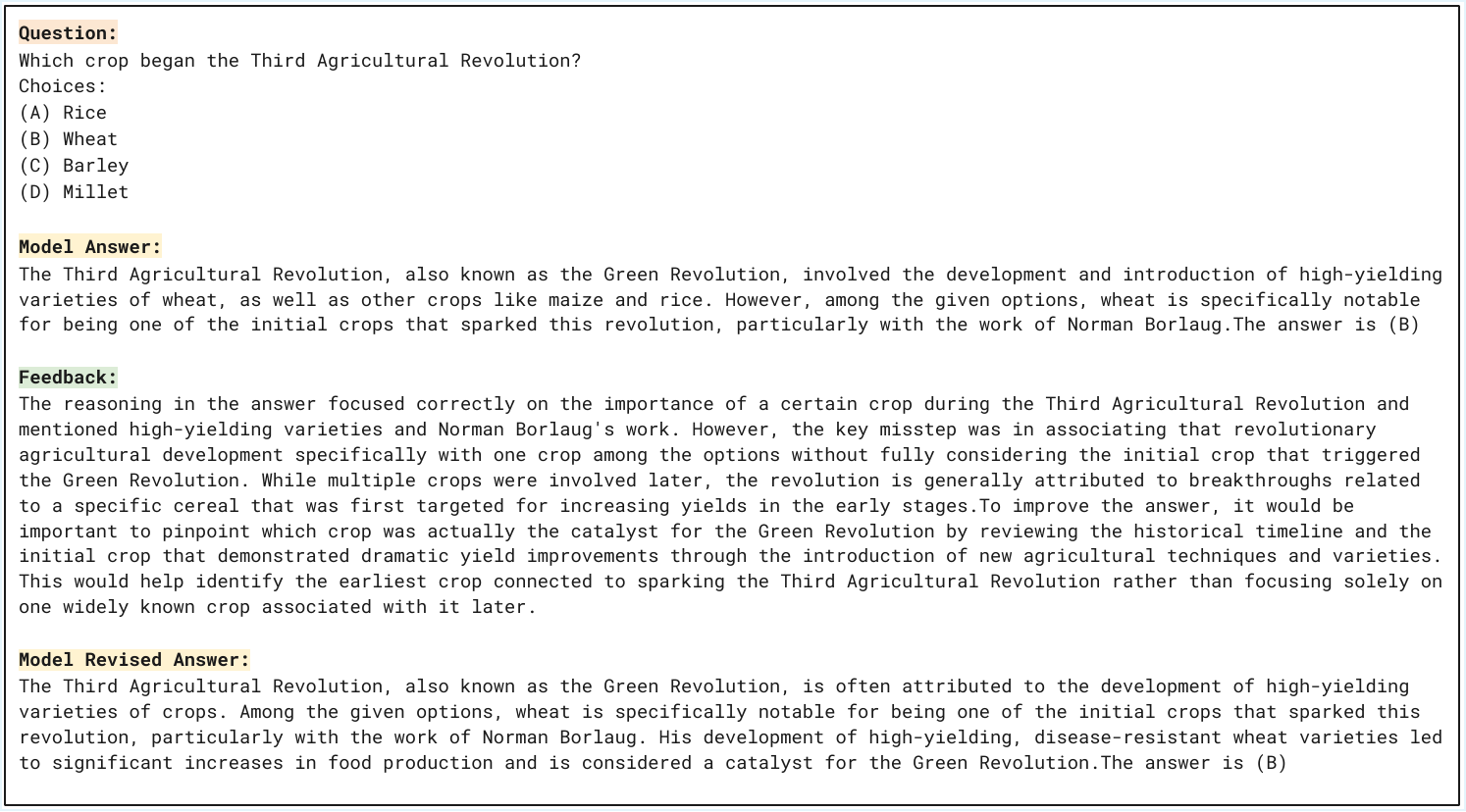}
    \caption{\llamafoursmallshort{} resisting feedback from \gpt{} in MMLU}
    \label{fig:enter-label}
\end{figure}

\begin{figure}[h!]
    \centering
    \includegraphics[width=1\linewidth]{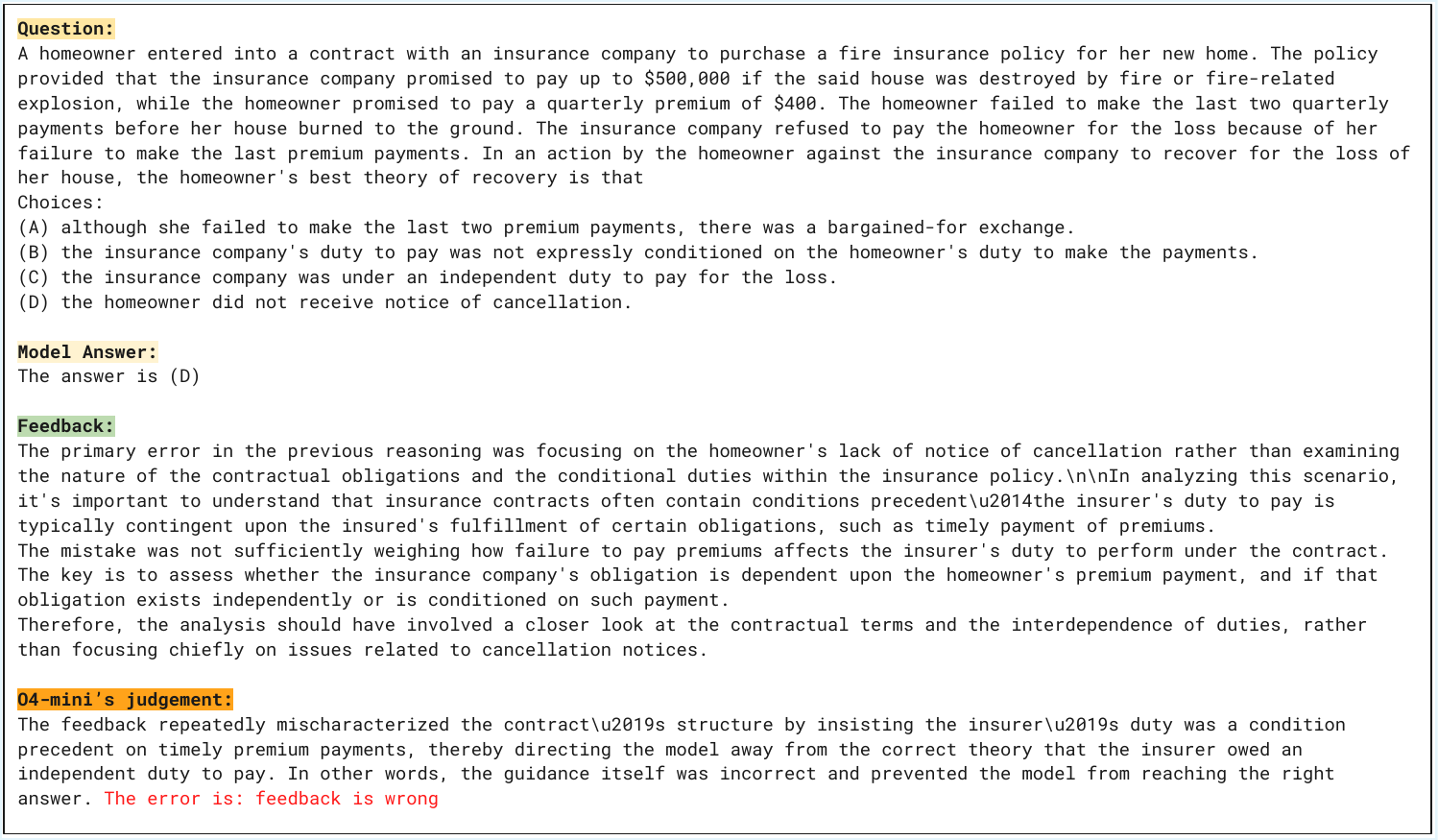}
    \caption{Wrong feedback provided by \gpt{} judged by o4-mini}
    \label{fig:wrongfeed}
\end{figure}

This section presents representative examples of persistent errors that prevent models from achieving correct solutions despite multiple feedback iterations. We illustrate the main error categories identified in our analysis: feedback resistance (where models fail to incorporate valid corrections, see \autoref{fig:enter-label}) and feedback quality issues (where the provided guidance is incorrect or misleading, see \autoref{fig:wrongfeed}).


\section{Synthetic digit multiplication task details} \label{appendix:synthetic_task}

\subsection{5-digit multiplication}

We construct a controlled arithmetic dataset consisting of 450 5-digit multiplication problems following the template: \texttt{``Calculate the following question: 19365 $\times$ 12534.''}

\begin{figure}[h!]
    \centering
    \includegraphics[width=1\linewidth]{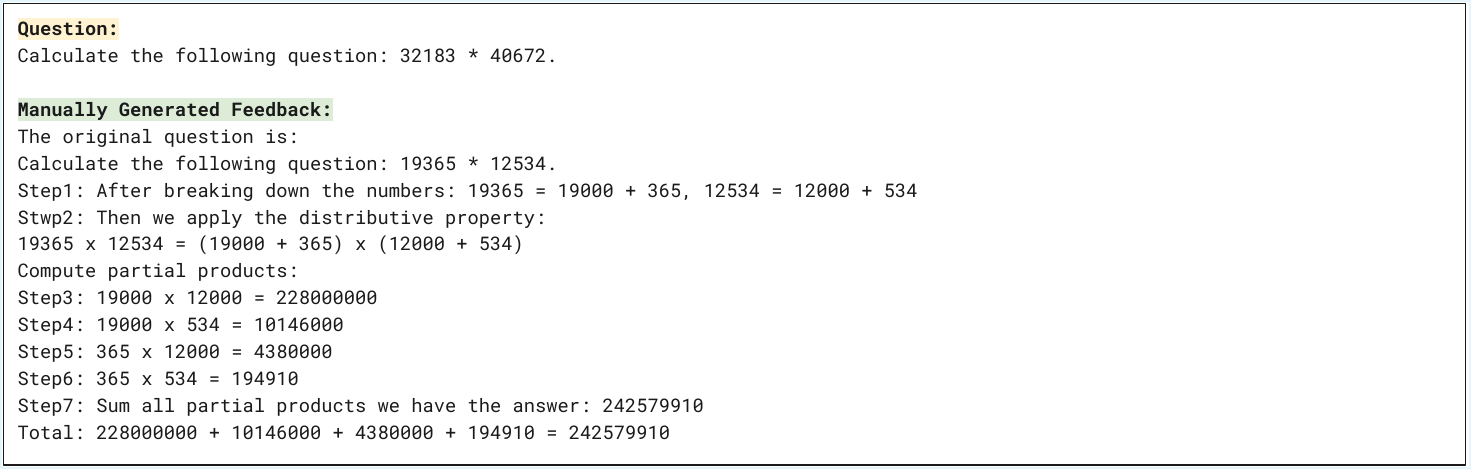}
    \caption{Templates for 5-digit multiplication solution.}
    \label{fig:5d-solution}
\end{figure}

\paragraph{Feedback generation} We employ a deterministic, human-designed template based on the distributive property to generate ground truth solutions. This template systematically decomposes each multiplication into partial products, providing a clear step-by-step solution pathway. Our template-based approach serves two key purposes: (1) demonstrating structured problem decomposition strategies for complex arithmetic, and (2) ensuring feedback correctness and interoperability. In \autoref{fig:5d-solution}, we illustrate an example template solution, which serves as the reference for feedback generation. The feedback model compares solver outputs against this structured breakdown to identify specific computational errors.

\paragraph{Answer masking strategy} To maintain task difficulty, we also mask intermediate partial answers before providing feedback to the solver model. 
The final feedback combines the masked template solution with model-generated guidance tailored to the specific errors observed.


\subsection{Hexadecimal 5-Digit Multiplication}

We also extend our synthetic arithmetic evaluation to hexadecimal multiplication, creating problems that challenge models' ability to work with non-standard number systems. The question template follows the format: \texttt{``Calculate the following question, where each number is represented in base 16: 69837 $\times$ 17635.''} All answers are expected in base-16 format.

\begin{figure}
    \centering
    \includegraphics[width=1\linewidth]{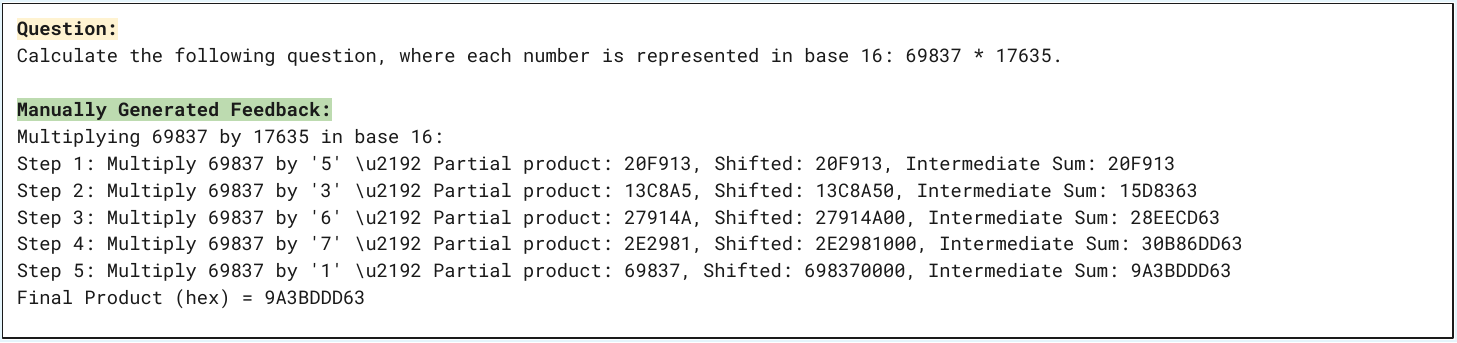}
    \caption{Hexadecimal 5-Digit Multiplication process solution.}
    \label{fig:hex-prompt}
\end{figure}

\paragraph{Template-based feedback.} Similar to decimal multiplication, we generate feedback using deterministic step-by-step templates. The multiplication process involves sequentially multiplying the first operand by each digit of the second operand (interpreting digits in base-16), then summing the resulting partial products with appropriate positional shifts. \autoref{fig:hex-prompt} demonstrates an example template solution. We validate solution correctness by verifying that partial product summation matches results from standard base-16 calculators.

\textbf{Masking strategy for hexadecimal multiplication.} Given the increased computational complexity of hexadecimal arithmetic, we employ a more permissive masking approach. We mask only the final summation step while preserving intermediate partial products. This design balances providing sufficient guidance with maintaining the core challenge of hexadecimal computation.

\section{Analysis of model confidence and \PHENOMENON{}} \label{appendix:confidence}

\begin{figure}[h!]
    \centering
    \begin{subfigure}[t]{0.48\linewidth}
        \includegraphics[width=\linewidth]{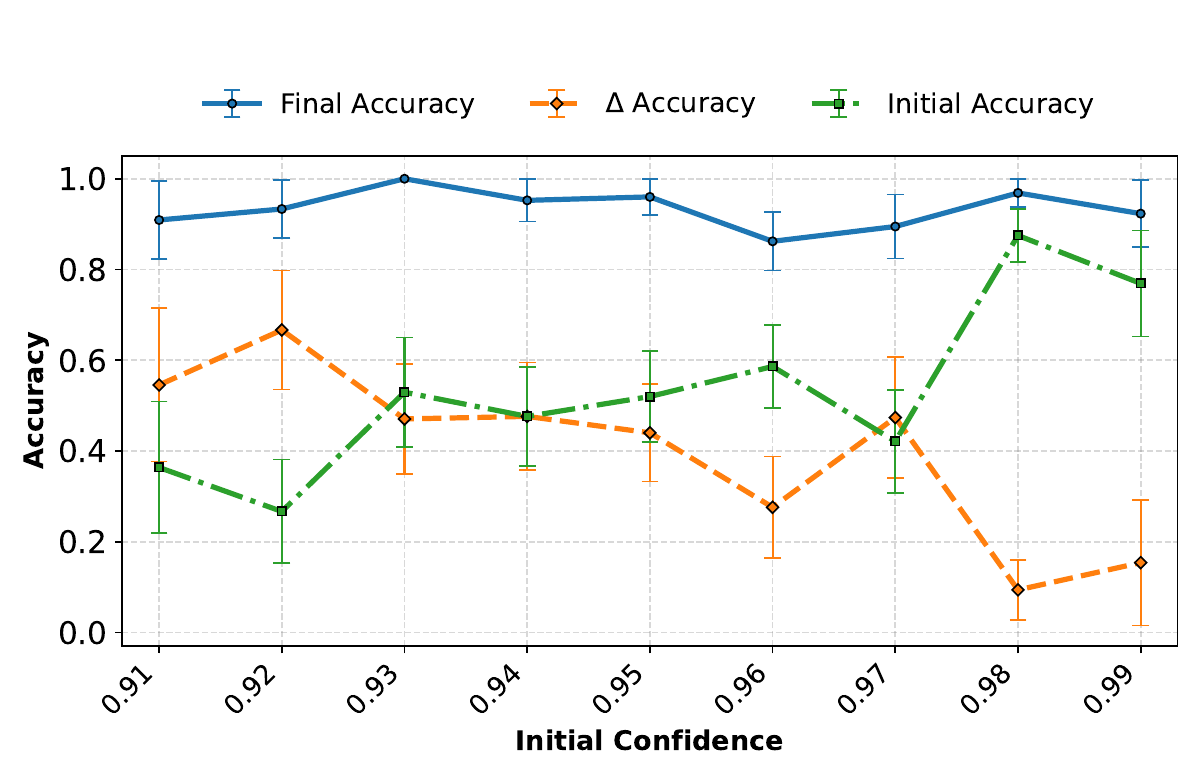}
        \caption{GPQA confidence vs. accuracy}
        \label{fig:scout_gpqa}
    \end{subfigure}
    \hfill
    \begin{subfigure}[t]{0.48\linewidth}
        \includegraphics[width=\linewidth]{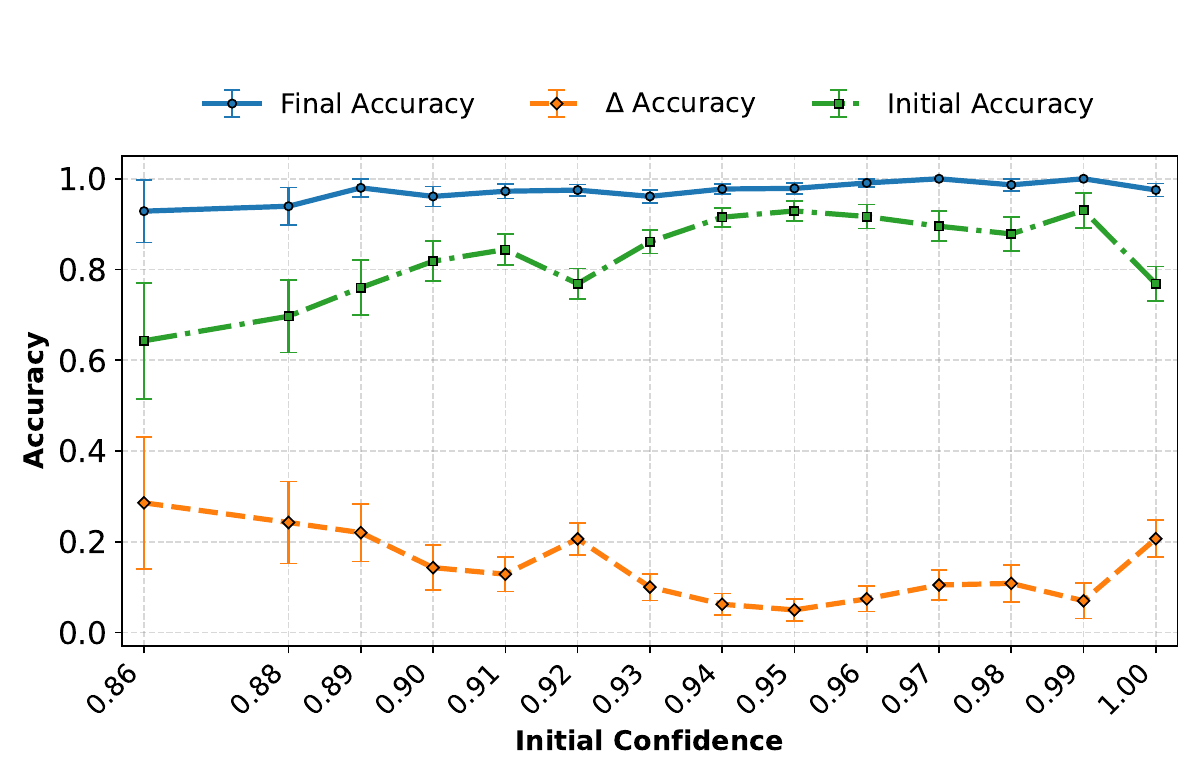}
        \caption{MMLU confidence vs. accuracy}
        \label{fig:scout_mmlu}
    \end{subfigure}

    \vspace{0.5em} 

    \begin{subfigure}[t]{0.48\linewidth}
        \includegraphics[width=\linewidth]{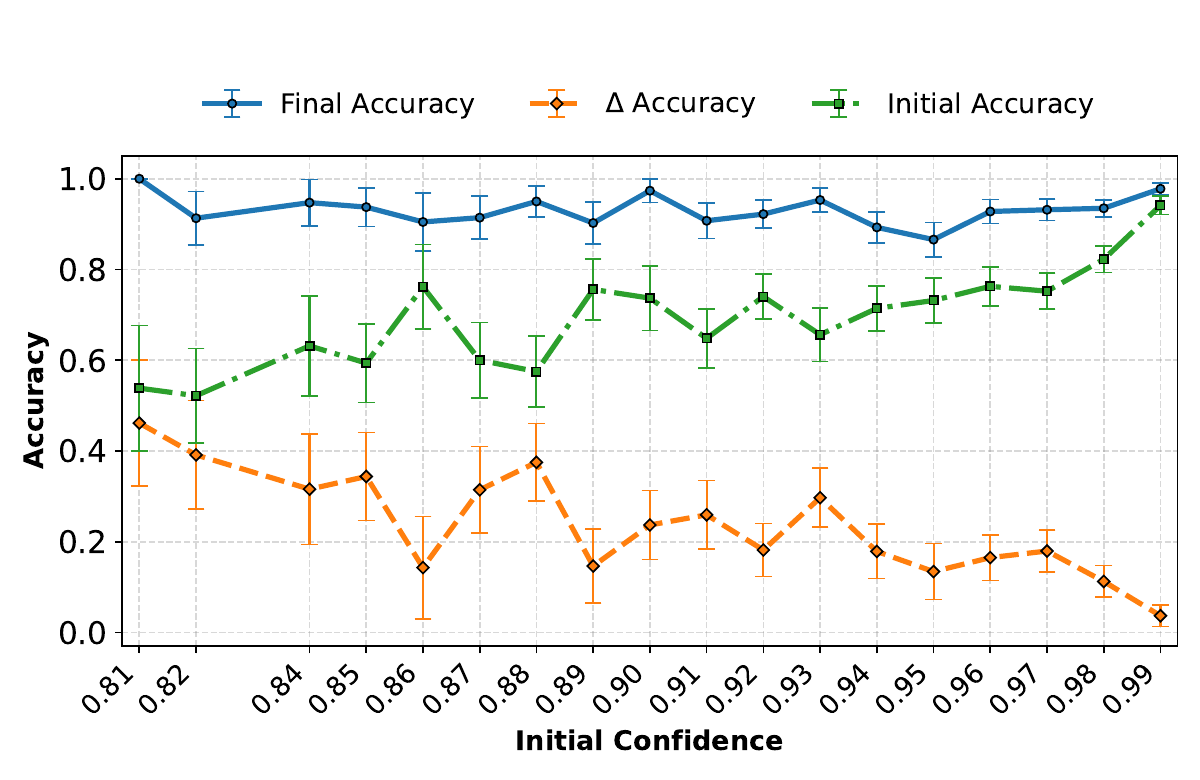}
        \caption{MMLU Pro confidence vs. accuracy}
        \label{fig:scout_mmlu_pro}
    \end{subfigure}
    \hfill
    \begin{subfigure}[t]{0.48\linewidth}
        \includegraphics[width=\linewidth]{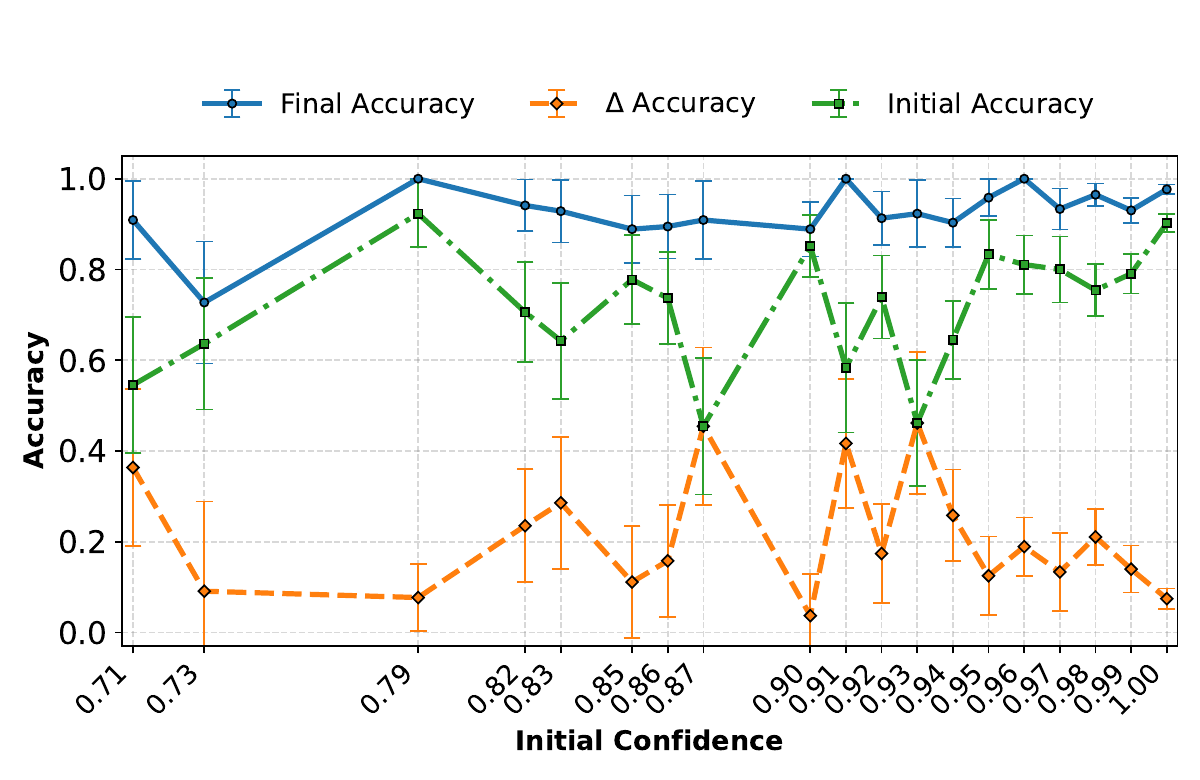}
        \caption{TriviaQA confidence vs. accuracy}
        \label{fig:scout_trivia}
    \end{subfigure}

    \caption{Confidence vs. accuracy across different datasets using \gpt{} as feedback model and \llamafoursmallshort{} as the solver.}
    \label{fig:conf_2x2_all}
\end{figure}

\autoref{fig:conf_2x2_all} presents confidence-accuracy relationships across four datasets: GPQA, MMLU, MMLU Pro, and TriviaQA. Each plot displays three metrics: initial accuracy at iteration 0, final accuracy after iterative feedback, and the improvement delta between them.

We define a model's \textit{initial confidence} in its generated answer as the exponential of the average log-probability per token in the answer sequence:

\[
\text{Initial Confidence} = \exp\left( \frac{1}{T} \sum_{t=1}^{T} \log p(a_t \mid a_{<t}, q) \right)
\]

where:
\begin{itemize}
    \item \( T \) is the number of tokens in the generated answer (with EOS excluded),
    \item \( a_t \) is the \( t \)-th token in the answer,
    \item \( a_{<t} \) denotes the prefix of the answer up to (but not including) position \( t \),
    \item \( q \) is the input question,
    \item \( p(a_t \mid a_{<t}, q) \) is the model's probability of generating token \( a_t \) given the previous tokens and the input.
\end{itemize}

This corresponds to the geometric mean of the per-token probabilities assigned by the model, and serves as a quantitative measure of the model's confidence in its complete answer.

We find that \textbf{initial confidence strongly predicts initial accuracy} across all datasets. Higher confidence bins consistently correspond to higher initial performance (Figures~\ref{fig:scout_gpqa}--\ref{fig:scout_trivia}), confirming that the model's confidence is a good predictor of its initial accuracy.

However, \textbf{confidence poorly predicts improvement potential}. The relationship between initial confidence and accuracy gains varies substantially:
\begin{itemize}
    \item GPQA shows peak improvements at moderate confidence levels, with diminishing returns at higher confidence
    \item MMLU and MMLU Pro exhibit relatively flat improvement patterns with the less confident questions getting more improvements. Nevertheless, this may caused by the low initial accuracy. 
    \item TriviaQA demonstrates erratic improvement fluctuations regardless of initial confidence
\end{itemize}


\section{Analysis of data familiarity and \textsc{Rigid Thinking}} \label{appendix:bias}

After analyzing data familiarity using answer frequency in the PopQA dataset, we found no clear correlation between model performance and frequency of answer words in the training data. However, surface-level frequency may not fully capture a model's true familiarity with content, as it fails to account for context quality, semantic relationships, and other factors affecting knowledge acquisition during pre-training. 

To better capture actual familiarity, we examine a more direct behavioral signal called in-domain performance: the model's accuracy in answering questions, measured using 100 generations per question with Llama-3.3. This behavioral familiarity metric reflects the cumulative effect of all factors contributing to the model's internalized knowledge.

\begin{figure}[h!]
  \centering
  \begin{subfigure}[b]{0.48\linewidth}
    \centering
    \includegraphics[width=\linewidth]{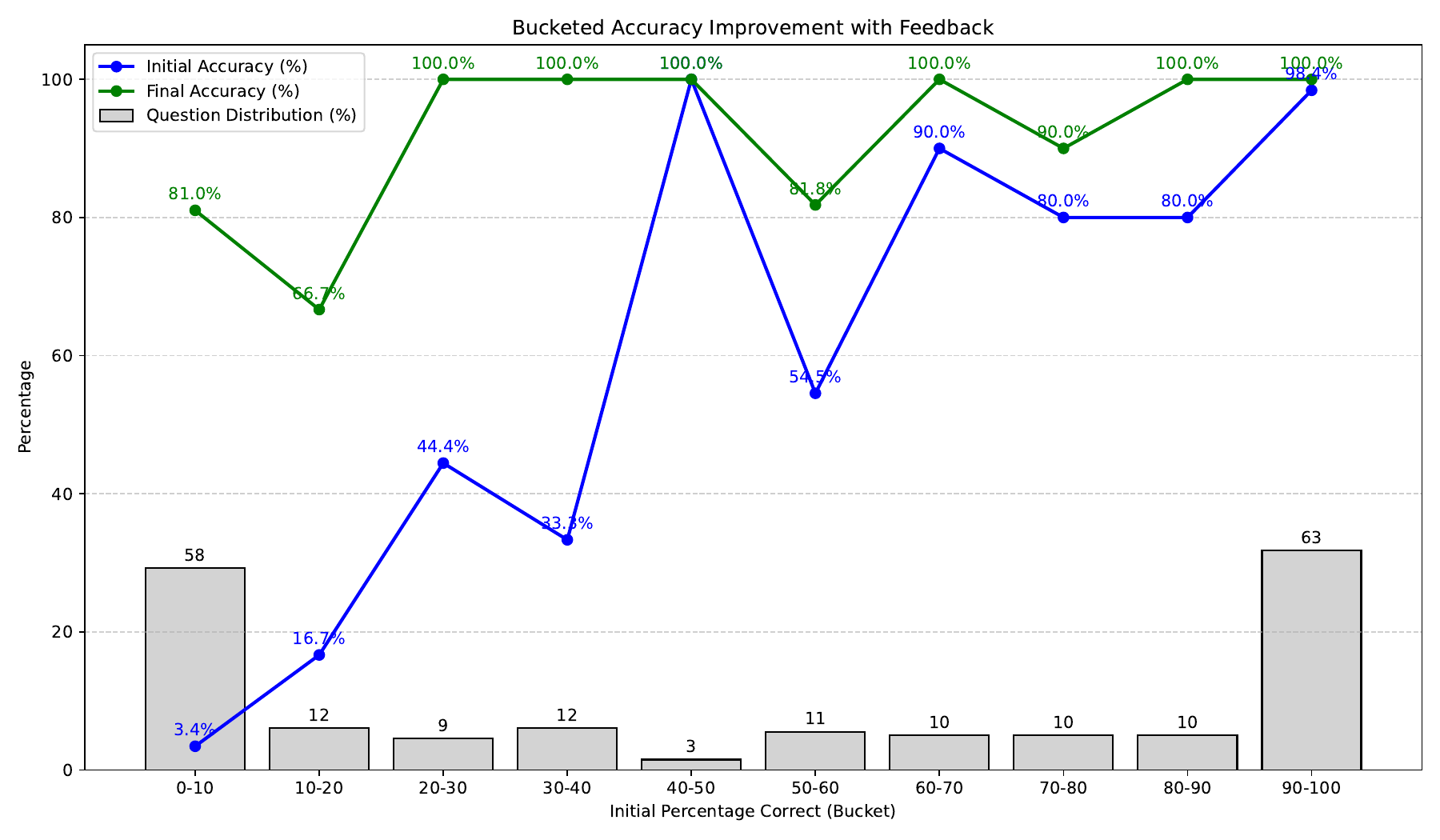}
    \caption{GPQA in-domain performance}
    \label{fig:gpqa-indomain}
  \end{subfigure}
  \hfill
  \begin{subfigure}[b]{0.48\linewidth}
    \centering
    \includegraphics[width=\linewidth]{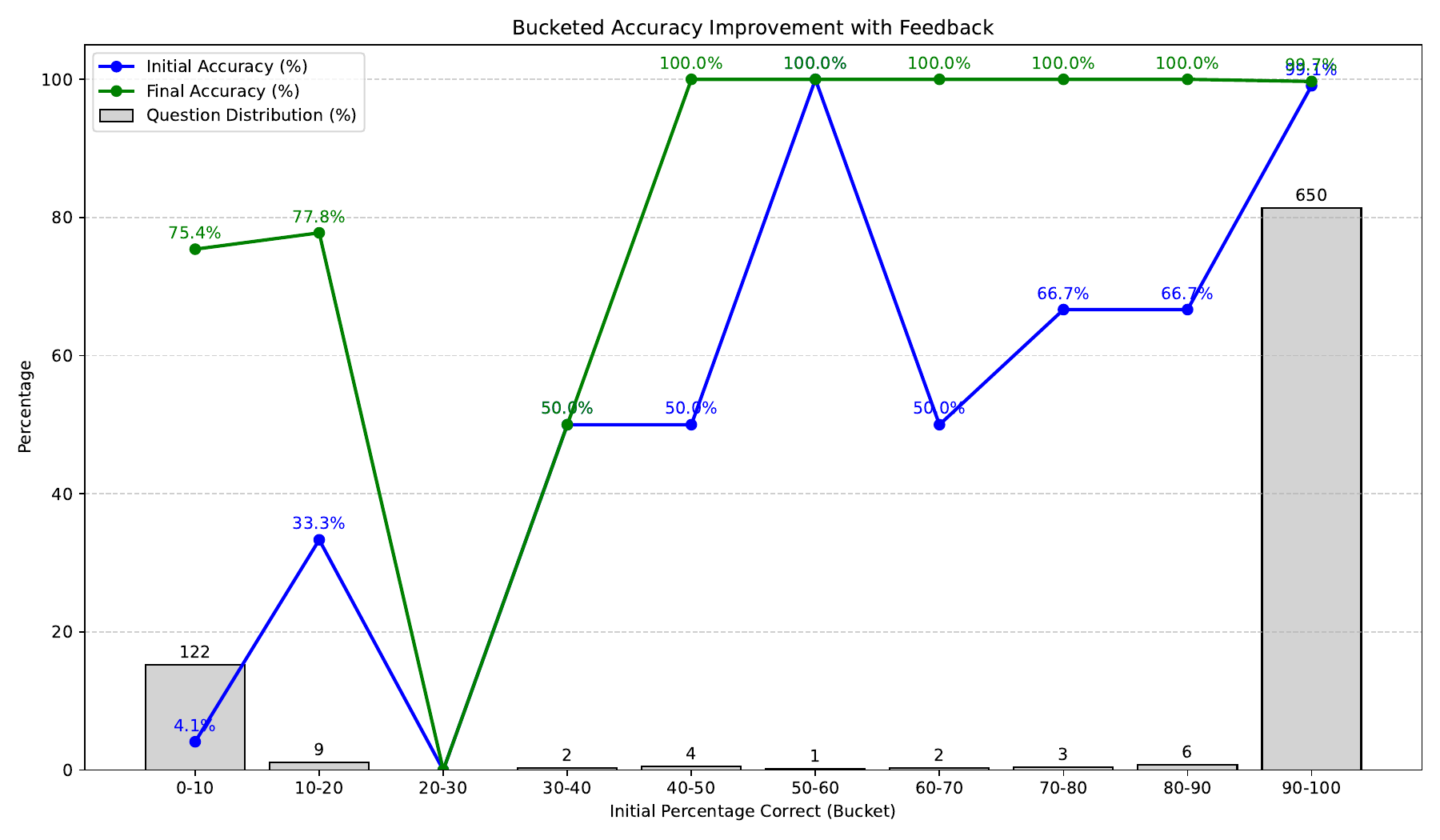}
    \caption{TriviaQA in-domain performance}
    \label{fig:triviaqa-indomain}
  \end{subfigure}

  \vspace{1em}

  \begin{subfigure}[b]{0.48\linewidth}
    \centering
    \includegraphics[width=\linewidth]{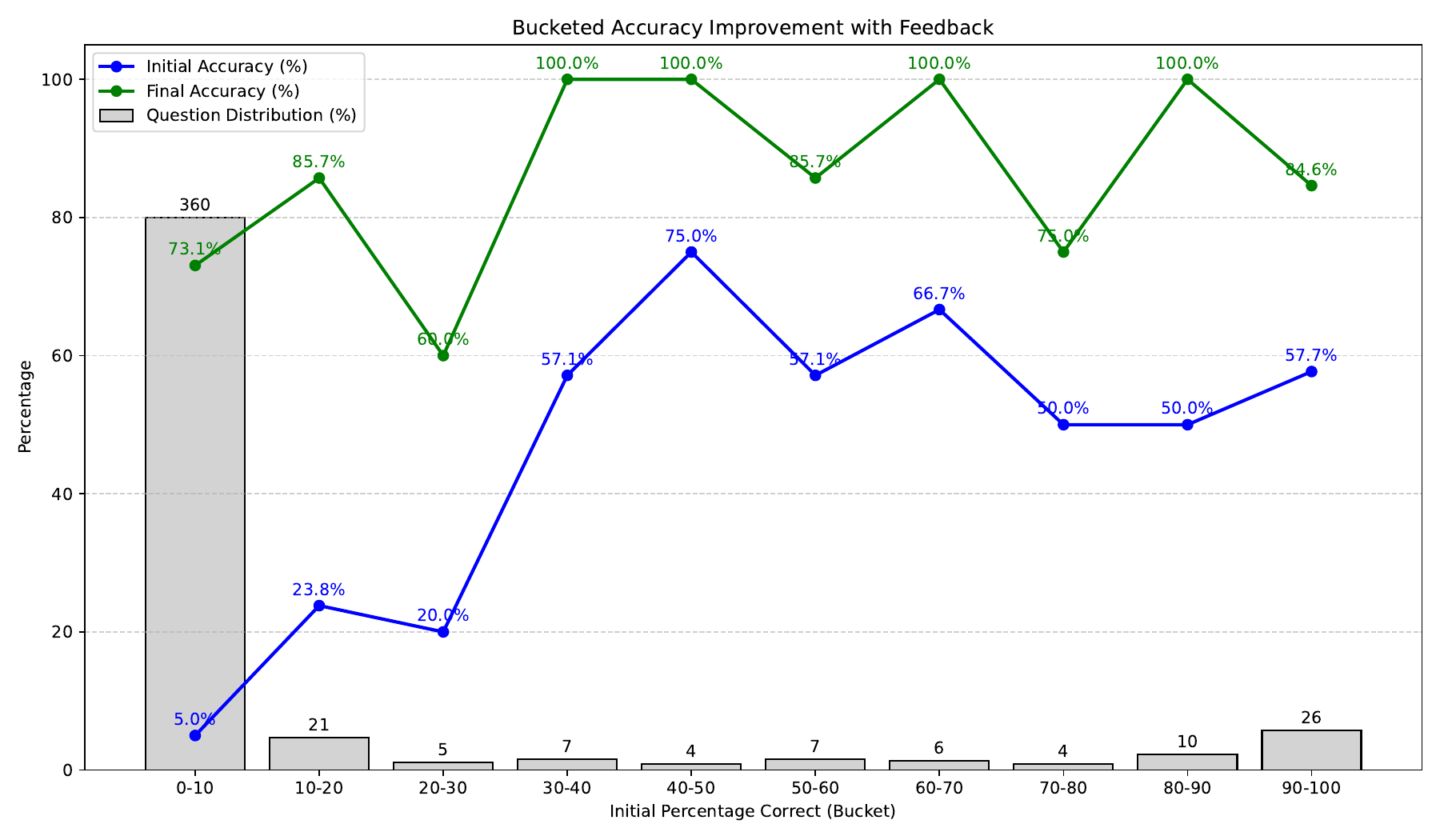}
    \caption{5-digit multiplication in-domain}
    \label{fig:5d-indomain}
  \end{subfigure}
  \hfill
  \begin{subfigure}[b]{0.48\linewidth}
    \centering
    \includegraphics[width=\linewidth]{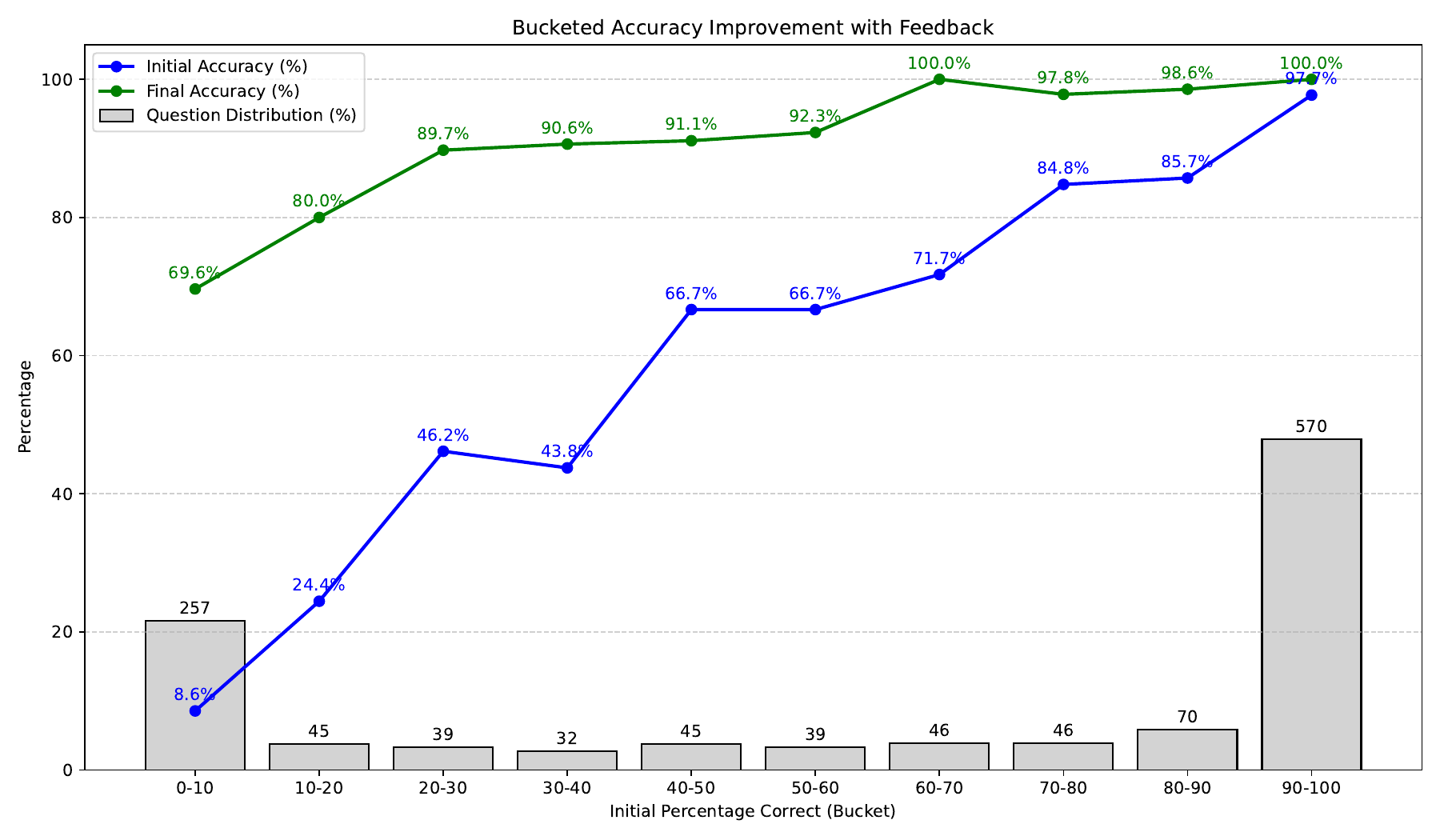}
    \caption{MMLU Pro in-domain performance}
    \label{fig:mmlupro-indomain}
  \end{subfigure}

  \caption{In-domain accuracy of \llamathreeshort{} across four benchmark tasks.}
  \label{fig:indomain-overview}
\end{figure}

\autoref{fig:indomain-overview} illustrates the in-domain performance of Llama-3.3 across four benchmarks—GPQA, TriviaQA, 5-digit multiplication, and MMLU Pro. We bucket questions based on the model's initial accuracy and report both initial and final accuracy after iterative feedback. While the model shows improvement across all buckets, questions with higher behavioral familiarity (higher initial accuracy) consistently achieve higher final accuracy as well. This suggests that behavioral familiarity is a more informative predictor of both current performance and improvement potential than answer frequency alone. Nevertheless, we still cannot obverse any consistent patterns across all these datasets in the initial vs. final accuracy. 

\section{Analysis of reasoning complexity and \PHENOMENON{}} \label{appendix:reasoning_complexity}
To investigate whether the model’s improvement over iterations correlates with question difficulty or the reasoning complexity, we compare the performance of \llamafoursmallshort{} on two synthesized multiplication tasks: 5-digit and 6-digit problems. Unlike prior datasets, which lack clear separability in difficulty levels, these tasks were manually constructed with 450 questions each to ensure a well-defined difference in complexity. 

The initial accuracy of \llamafoursmallshort{} is 2.2\% on 5-digit multiplication and 0.889\% on 6-digit multiplication. Interestingly, while the 6-digit task is objectively more difficult, we observe greater improvement across iterations compared to the 5-digit task. One possible explanation is that more difficult tasks offer more room for feedback-driven correction because solver model has less initial knowledge about how to solve them. However, we also observe cases where simpler questions yield higher final accuracy, suggesting that the relationship between task complexity and feedback effectiveness is non-monotonic and influenced by additional factors.
\begin{figure}[h!]
    \centering
    \includegraphics[width=0.8\linewidth]{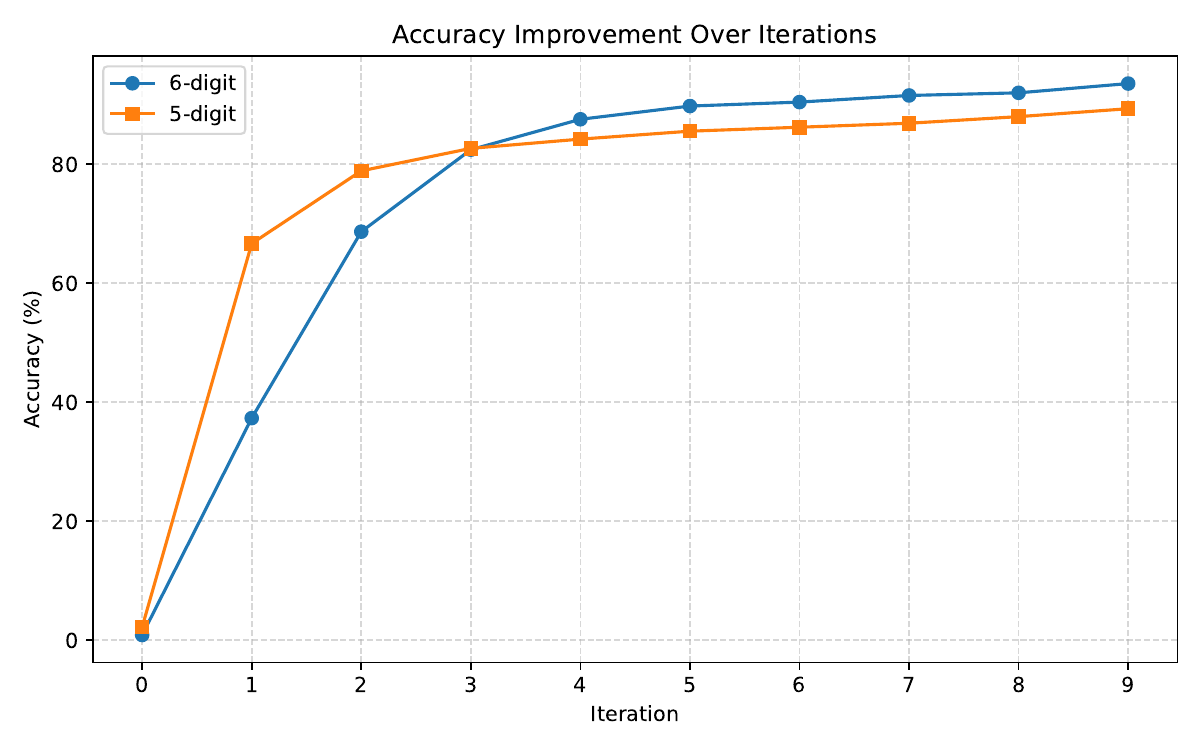}
    \caption{Comparison of the Improvement for 5-digit and 6-digit multiplication with \gpt{} as feedback model}
    \label{fig:comp_56}
\end{figure}

\section{Analysis of model type and \PHENOMENON{}} \label{appendix:model_type}

To better understand the overlap in failure cases among \llamathreeshort{}, \llamafoursmallshort{}, and \llamafourbigshort{}, we compared their incorrect predictions across several benchmark datasets. Specifically, we report the number of shared mistakes between each pair of models, the number of questions all three models got wrong, the total number of unique mistakes (union), and the \textbf{Overlap Ratio}, defined as the proportion of all-three common errors to the total number of distinct errors.

The \textit{Overlap Ratio} offers a normalized measure of agreement in model failures. Notably, \textbf{AIME} shows the highest overlap (35.7\%), suggesting a subset of examples that all three models consistently struggle with. Conversely, datasets such as \textbf{GPQA} and \textbf{5-digit Multiplication} exhibit minimal overlap (6.9\% and 0.7\%, respectively), indicating that the models tend to fail on different questions.

These findings suggest that model failures are often idiosyncratic rather than being concentrated around a universally difficult subset of examples. The relatively low overlap across datasets highlights the challenge of achieving robust self-correction: errors are not easily attributable to a common set of pitfalls, but rather reflect distinct weaknesses in each model’s reasoning and generalization.

\begin{table}[h!]
\centering
\scriptsize
\resizebox{\textwidth}{!}{%
\begin{tabular}{lcccccc}
\toprule
\textbf{Dataset} & \textbf{L3.3–Scout} & \textbf{L3.3–Maverick} & \textbf{Scout–Maverick} & \textbf{All-Three} & \textbf{Union} & \textbf{Overlap Ratio} \\
\midrule
AIME        & 8  & 9  & 5  & 5  & 14  & 0.357 \\
TriviaQA             & 22 & 16 & 28 & 14 & 105 & 0.133 \\
MATH-500         & 18 & 14 & 11 & 9  & 64  & 0.141 \\
MMLU             & 15 & 12 & 19 & 11 & 55  & 0.200 \\
MMLU Pro         & 49 & 40 & 43 & 30 & 163 & 0.184 \\
GPQA             & 3  & 5  & 2  & 2  & 29  & 0.069 \\
5-digit Mult.    & 21 & 3  & 1  & 1  & 141 & 0.007 \\
\bottomrule
\end{tabular}
}
\caption{Pairwise and three-way common failure cases among \llamathreeshort{}, \llamafoursmallshort{}, and \llamafourbigshort{} across datasets. Overlap Ratio is computed as the number of questions all three models failed on divided by the union of all distinct failures.}
\label{tab:three_model_pitfalls}
\end{table}

\section{Details of testing model's self-perception of \PHENOMENON{}} \label{appendix:example_model_self_perception}
\begin{figure}[h!]
    \centering
    \includegraphics[trim=0.0cm 10.0cm 0cm 0cm,width=1\linewidth]{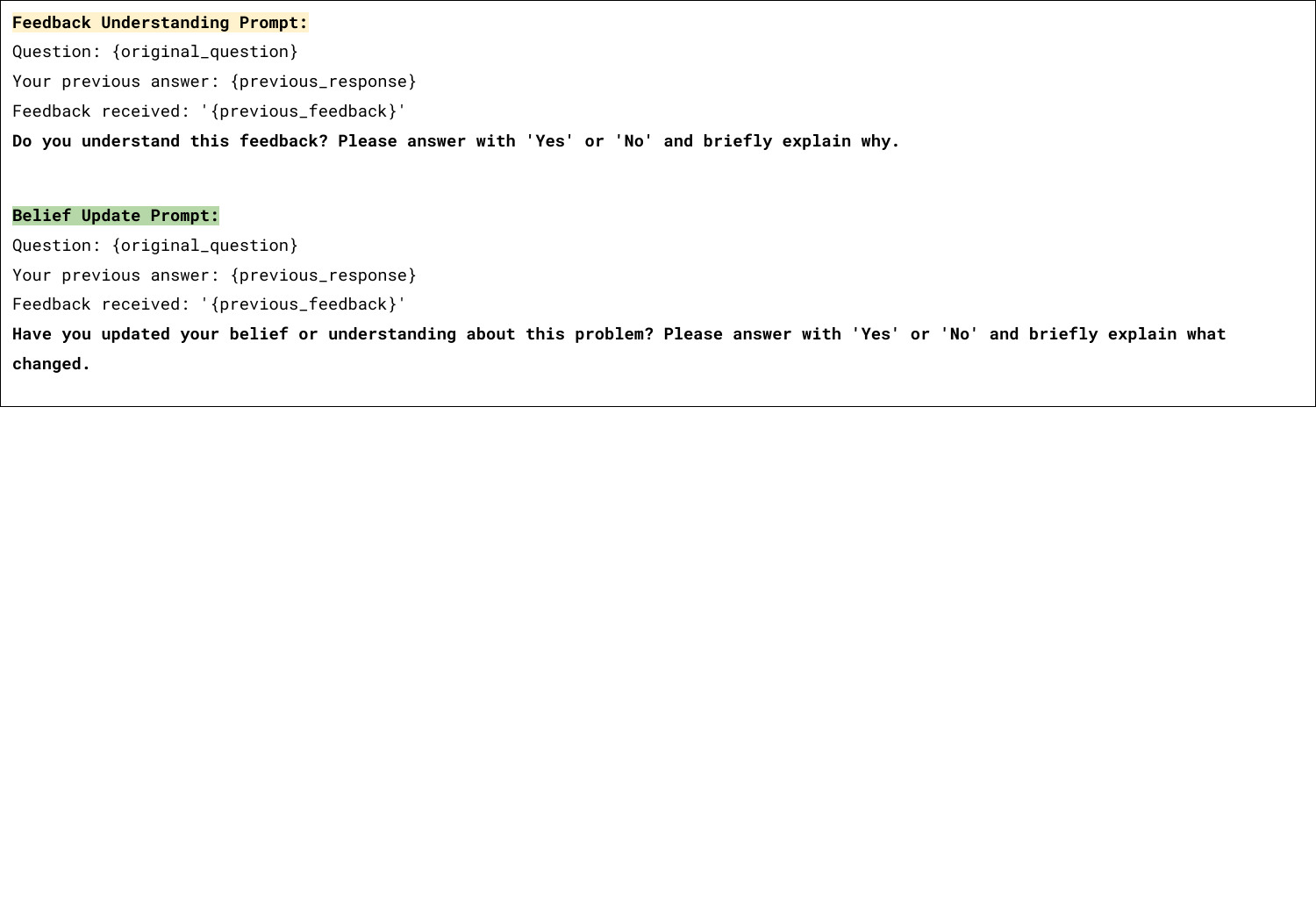}
    \caption{An example of \llamafoursmallshort{} explicitly acknowledging changing its belief while still making the same mistake in the next attempt.}
    \label{fig:belief_probing_prompt}
\end{figure}

\autoref{fig:belief_probing_prompt} shows the prompts we used to probe whether the model understand the given feedback and whether it's willing to change its belief.

\begin{figure}[h!]
    \centering
    \includegraphics[trim=0.0cm 1.0cm 0cm 0cm,width=1\linewidth]{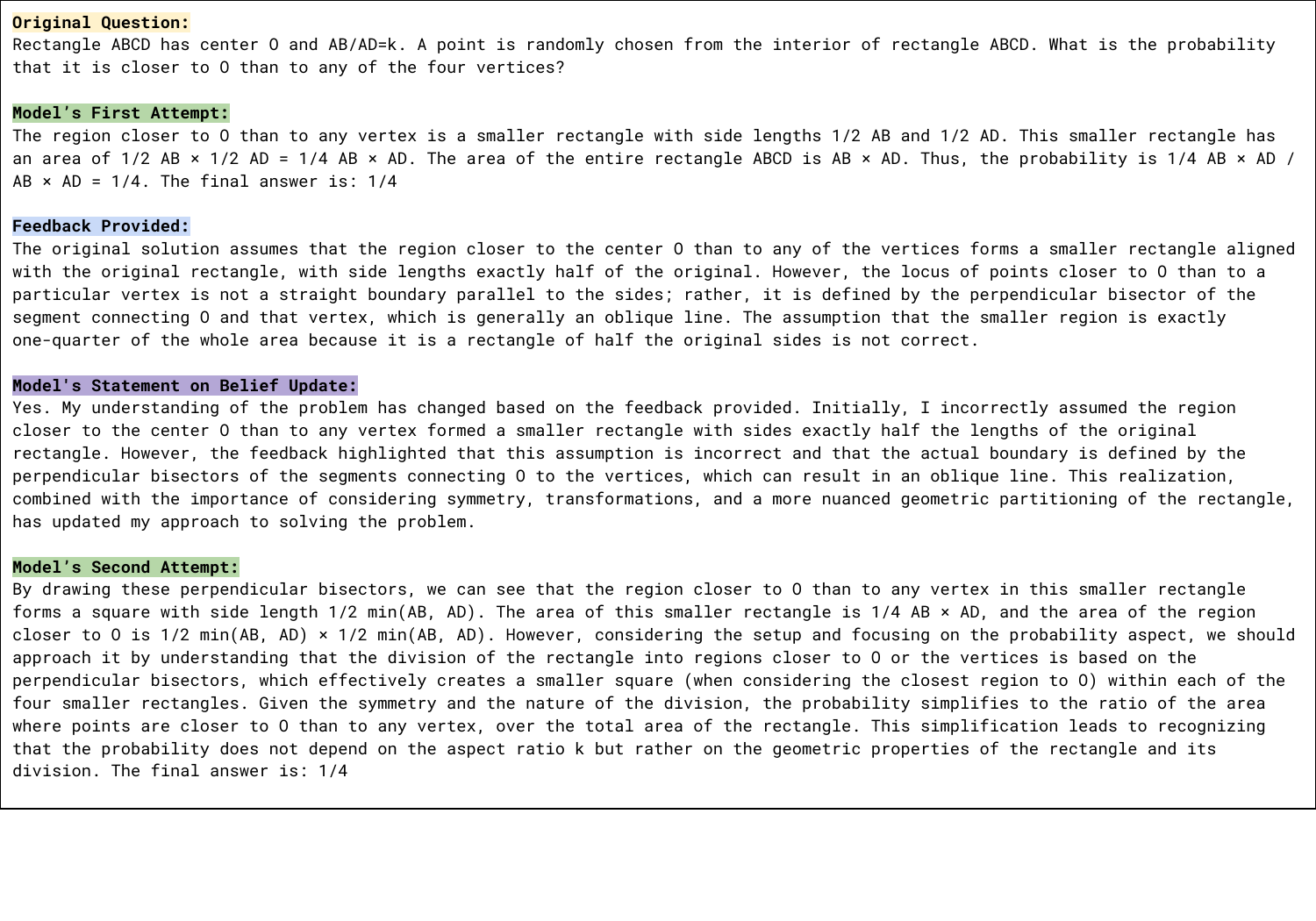}
    \caption{An example of \llamafoursmallshort{} explicitly acknowledging changing its belief while still making the same mistake in the next attempt.}
    \label{fig:example_self_indentification}
\end{figure}

\autoref{fig:example_self_indentification} shows an example where \llamafoursmallshort{} explicitly acknowledges that ``the actual boundary is defined by the perpendicular bisectors of the segments connecting O to the vertices, which can result in an oblique line'' and claims its approach has been updated. Yet in its second attempt, it still describes the region as forming ``a square with side length 1/2 min(AB, AD)'' and arrives at the same incorrect answer of 1/4. This demonstrates the model can articulate understanding of the feedback while its actual problem-solving behavior remains unchanged. The correct answer is 1/2.

\section{Full results of semantic entropy} \label{appendix:semantic_entropy}
\begin{figure}[h!]
    \centering
    \includegraphics[width=1\linewidth]{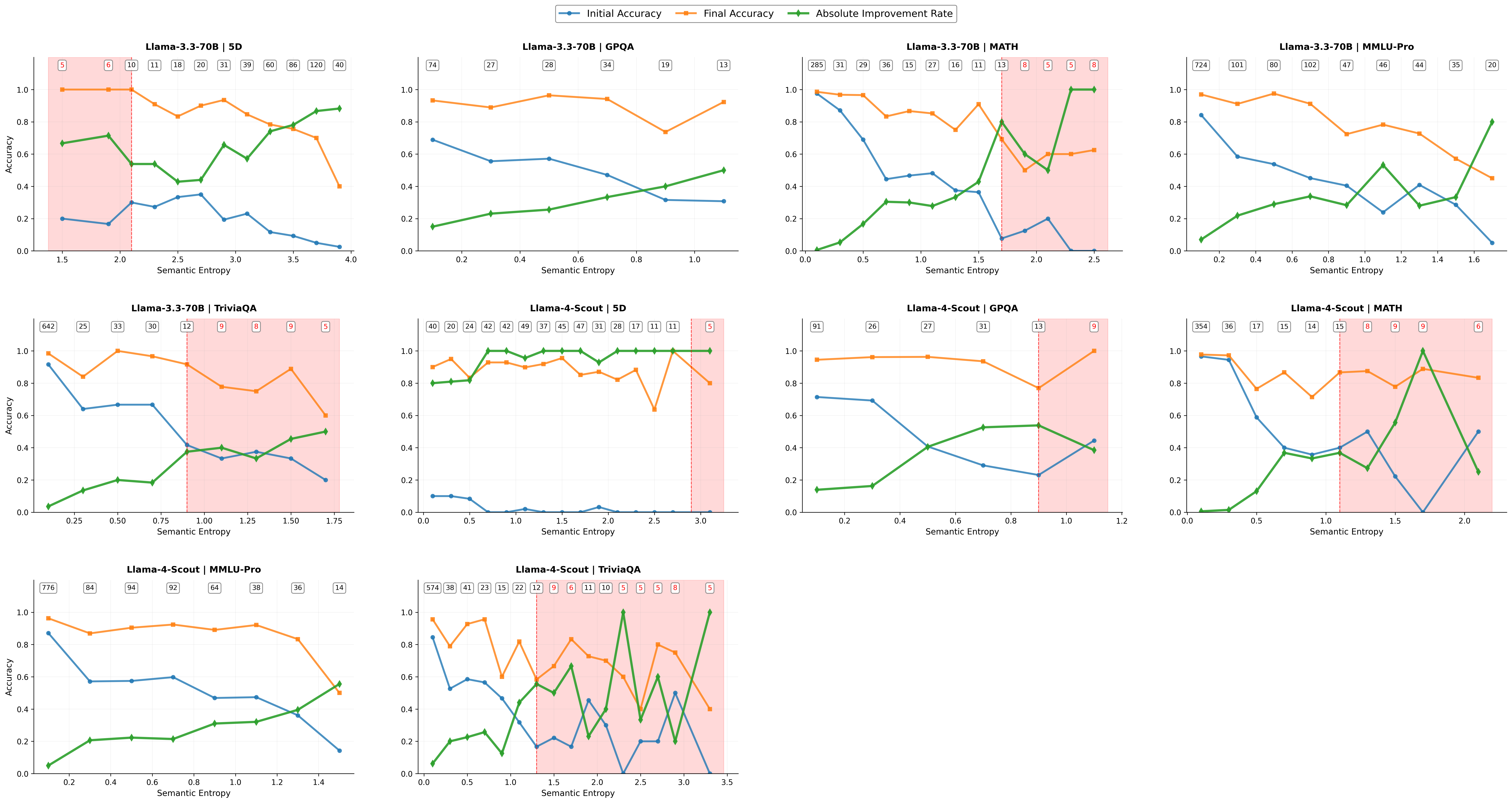}
    \caption{Full results of semantic entropy over five benchmark tasks and two models.}
    \label{fig:full_semantic_entropy_results}
\end{figure}

\autoref{fig:full_semantic_entropy_results} presents the complete self-correction analysis across all five benchmark tasks (5 digits multiplication, GPQA, MATH, MMLU-Pro, and TriviaQA) for both Llama-3.3-70B and Llama-4-Scout. Each subplot shows initial accuracy (blue), final accuracy (orange), and absolute improvement rate (green) as functions of semantic entropy with bucket size 0.2. Regions with fewer than 10 samples are marked with red shading and dashed boundaries to indicate unreliable data, with sample counts displayed in boxes above each plot.

\end{document}